\documentclass[twocolumn]{svjour3} 
\smartqed
\usepackage{graphicx}
\usepackage{subfigure}
\usepackage{booktabs}
\usepackage{multirow}
\usepackage{amssymb}
\usepackage{color}
\usepackage{natbib}
\usepackage[colorlinks, linkcolor=red, anchorcolor=red, citecolor=blue]{hyperref}
\usepackage{pifont}
\usepackage{textcomp}

\begin{document}

\title{DarkVision: A Benchmark for Low-light Image/Video Perception}






\author{Bo~Zhang$^1$\and Yuchen~Guo$^{2,3}$\and Runzhao~Yang$^1$\and Zhihong~Zhang$^1$\and Jiayi~Xie\and Jinli~Suo$^{1,3}$\and Qionghai~Dai$^{1,2,3}$}

\institute{
Jinli Suo\at
jlsuo@tsinghua.edu.cn
\and
Qionghai Dai\at
qhdai@tsinghua.edu.cn
\and
Bo Zhang \at b-zhang18@mails.tsinghua.edu.cn
\and
Yuchen Guo\at yuchen.w.guo@gmail.com
\and
Runzhao Yang\at yangrz20@mails.tsinghua.edu.cn
\and
Zhihong Zhang\at zhangzh19@mails.tsinghua.edu.cn
\and
Jiayi Xie\at xiejiayi97@163.com
\and
$^1$ Department of Automation, Tsinghua University, Beijing, China\at 
$^2$ Beijing National Research Center for Information Science and Technology (BNIST)\at
$^3$ Institute for Brain and Cognitive Sciences, Tsinghua University (THU-IBCS) 
}

\date{Received: date / Accepted: date}

\maketitle

\begin{abstract}
Imaging and perception in photon-limited scenarios is necessary for various applications, e.g., night surveillance or photography, high-speed photography, and autonomous driving. In these cases, cameras suffer from low signal-to-noise ratio, which degrades the image quality severely and poses challenges for downstream high-level vision tasks like object detection and recognition. Data-driven methods have achieved enormous success in both image restoration and high-level vision tasks. However, the lack of high-quality benchmark dataset with task-specific accurate annotations for photon-limited images/videos delays the research progress heavily. In this paper, we contribute the first multi-illuminance, multi-camera, and low-light dataset, named \textit{DarkVision}, serving for both image enhancement and object detection.
We provide bright and dark pairs with pixel-wise registration, in which the bright counterpart provides reliable reference for restoration and annotation. 
The dataset consists of bright-dark pairs of 900 static scenes with objects from 15 categories, and 32 dynamic scenes with 4-category objects. For each scene, images/videos were captured at 5 illuminance levels using three cameras of different grades, and average photons can be reliably estimated from the calibration data for quantitative studies. The static-scene images and dynamic videos respectively contain around 7,344 and 320,667 instances in total.
With DarkVision, we established baselines for image/video enhancement and object detection by representative algorithms. { To demonstrate an exemplary application of DarkVision, we propose two simple yet effective approaches for improving performance in video enhancement and object detection respectively.} 
We believe DarkVision would advance the state-of-the-arts in both imaging and related computer vision tasks in low-light environment. Our dataset will be released online for academic use after paper publication.
\end{abstract}




\section{Introduction}\label{sec:introduction}

Imaging with insufficient photons arriving at the sensor, referred to as \textbf{low-light imaging}, happens when there is weak ambient light irradiance or short exposure time. Low-light imaging is inevitable in both daily-life applications and scientific research, such as nighttime photography, surveillance and autonomous driving, as well as high-speed and transient imaging. It also has wide applications in microscopic cases such as fluorescence imaging.


The low-light imaging and analysis are heavily challenged by the photon starvation problem which significantly degrades the image quality causing low signal-to-noise ratio (SNR) and makes downstream high-level tasks like object detection/tracking and human pose estimation more difficult. Many research efforts have been made to address this issue, which can be roughly categorized into two groups. The first group is based on specific hardwares, such as invisible-light sensors like infrared camera, X-ray imaging and Terahertz imaging equipment. Shin et al.~\citep{shin2016photon} developed a photon-efficient imaging setup with a single photon avalanche diode (SPAD) array to achieve imaging under extremely weak environment with averagely 1 detected signal photon for each pixel. Morris et al.~\citep{morris2015imaging} adopted the emerging ghost-imaging configuration for photography with a smaller number of photons with an average of fewer than one detected photon per pixel. Hasinoff et al.~\citep{hasinoff2016burst} proposed a computational photography pipeline that captures, aligns, and merges a burst of frames to reduce noise and increase dynamic range on mobile cameras. In spite of their effectiveness, these techniques or equipments cannot be widely applicable for consumer-grade imaging system due to high hardware cost, expertise-demanding usage and poor portability.
The second group, on the other hand, focuses on novel algorithms specially designed for low-light illuminations. Since algorithms and software are less expensive and easier for use and distribution, they have attracted considerable research interest in the recent decade~\citep{dong2011fast,hu2014deblurring,li2018structure,remez2017deep,park2017low}, which is also the focus of this paper.

During these years, data-driven approaches have been emerging and flourishing~\citep{8305143,lore2017llnet,Chen_2018_CVPR,li2018lightennet}, with promising performance on low-light image enhancement. With the image prior extracted from the large-scale dataset, these methods present excellent adaptiveness and effectiveness in this extremely challenging realm.
In the meantime, a number of low-light datasets have been collected for low-light data analysis~\citep{LOH201930,Wang_2019_CVPR,8259342,shen2017msr,5557884,Chen_2018_CVPR,liubenchmarking,10.1145/2713168.2713194}, either by numerical synthesis or real capturing. These timely benchmarks fire up the rapid emergence of a number of low-light image enhancement methods. Nevertheless, there are still many difficult but useful questions that are rarely explored in low-light imaging such as low-light video enhancement and object detection. The biggest obstacle is the fact that most existing low-light datasets are oriented to static image enhancement and with no or only limited annotations of single object class, leaving the analysis for dynamic videos and complicated high-level vision tasks impossible. Moreover, during the imaging process in low-light scenarios, limited photons are highly corrupted by various types of noise and usually result in a visually dark image. But what degree of 'darkness' do these images belong to? How would the darkness level of these datasets affect the enhancement difficulty? What is the boundary of current methods? Unfortunately, such intriguing questions are less probed not because they are not important but existing low-light datasets have little or ambiguous description for the darkness level of their data, which is insufficient to support such and further research.

\begin{figure*}[t]
    \vspace{-2mm}
	\centering
	\includegraphics[width=\linewidth]{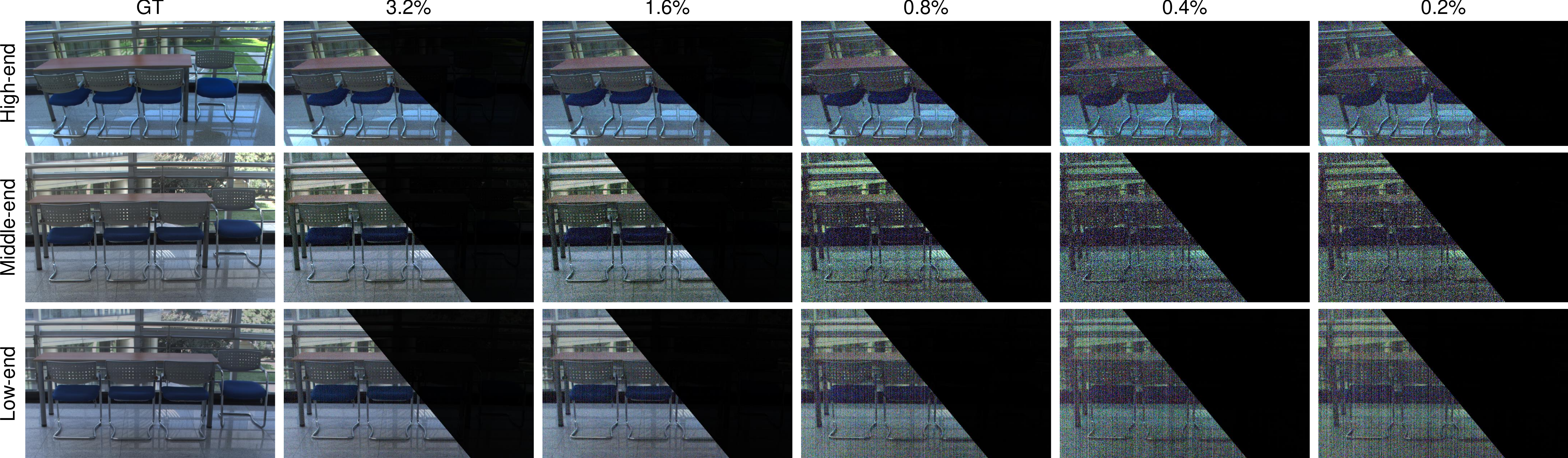}
	\caption{A typical example in the static subset of DarkVision. For each scene, we provide the bright reference and its dark counterparts at 5 illuminance levels. Here we linearly scale the dark images towards the bright reference to demonstrate the degeneration as illumination level decreases. We also apply Gamma transformation to all the images for better visualization.}
	\label{fig:ExampleImg}
\end{figure*}

\begin{figure*}[t]
	\centering
	\includegraphics[width=\linewidth]{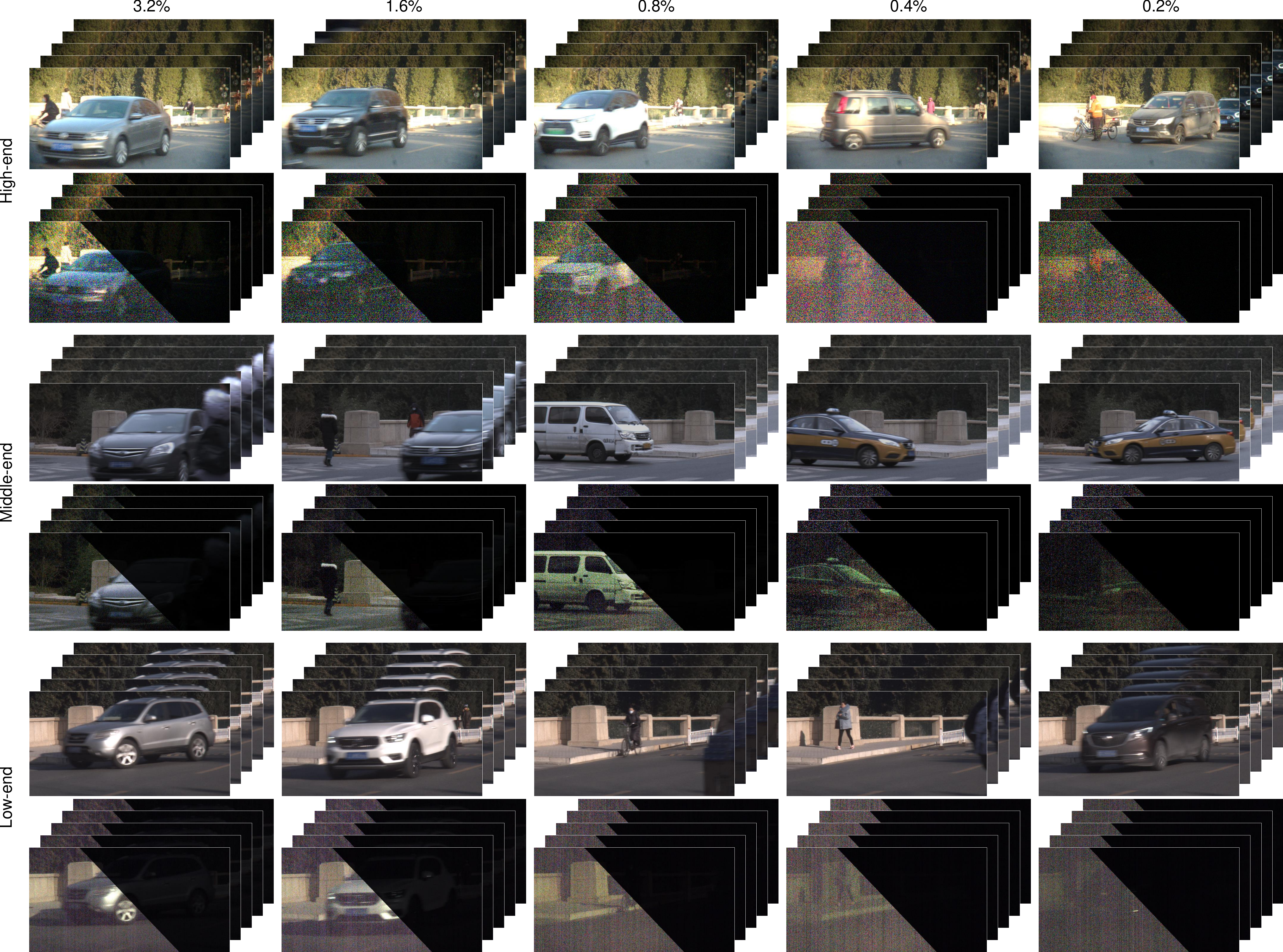}
	\caption{A typical example in the dynamic subset of DarkVision. We apply the same enhancement as in Fig.~\ref{fig:ExampleImg} for better visualization. Note that the foreground at different illuminance levels is not identical, although captured at the same location.}
	\vspace{-2mm}
	\label{fig:ExampleVideo}
\end{figure*}

\begin{table*}
	\centering
	\caption{{\color{black}Comparison between DarkVision and other low-light datasets. Here ``---" means that the data is not available. BD denotes bit depth. R and S denote real and synthetic respectively.}}\vspace{-2mm}
	\begin{tabular}{lllllllc}
           \midrule
            \textbf{Datasets} & Type & $\#$ Scenes & $\#$ Images & BD & Annotations  & R/S & Paired? \\
            \toprule
            LLNet\citep{lore2017llnet} & Image & 169 & 169 & 8 & No & S & \checkmark\\
            MSR-Net\citep{shen2017msr} & Image & 10,000 & 10,000 & --- & No & S & \checkmark \\
            SID\citep{Chen_2018_CVPR} & Image & 424 & 5,094 & 14& No & R & \checkmark \\
            SICE\citep{8259342} & Image & 589 & 4,413 & 8 & No & R & \checkmark\\
            RENOIR\citep{anaya2018renoir} & Image & 120 & 1,500& 12 & No & R & \checkmark \\
            LOL\citep{Chen2018Retinex} & Image & 500 & 500&  8 & No & R+S & \checkmark\\
            DeepUPE\citep{Wang_2019_CVPR} & Image & 3,000 & 3,000&  --- & No & R+S & \checkmark\\
            VE-LOL-L\citep{liubenchmarking} & Image & 2,500 & 2,500 & 8 & No & R+S & \checkmark\\
            AGLLNet~\citep{lv2021attention} & Image & 22,656 & 22,656 & 8 & No & S & \checkmark \\
            SMOID\citep{Jiang_2019_ICCV} & Video & 179 & 35,800 &  12 & No & R & \checkmark \\
            DRV\citep{Chen_2019_ICCV} & Video & 202 & 22,220 & 14 & No & R & \checkmark \\
            SDSD\citep{9710730} & Video & 150 & 25,625 & 8 & No & R & \checkmark \\
            ExDARK\citep{LOH201930} & Image & 7,363 & 7,363 & 8 & 12-class boudning boxes & R & \ding{53} \\
            VE-LOL-H\citep{liubenchmarking} & Image & 10,940 & 10,940 & 8 & Face boudning boxes & R & \ding{53}\\
            Nighttime Driving & \multirow{2}{*}{Video} & 50 & 50 & \multirow{2}{*}{8} & 19-class semantic labels & \multirow{2}{*}{R} & \multirow{2}{*}{\ding{53}}\\
            \citep{daytime:2:nighttime} & & --- & 35,000 & & No semantic labels &  & \\
            Dark Zurich & \multirow{2}{*}{Video} & 151 & 151 & \multirow{2}{*}{8} & 19-class semantic labels & \multirow{2}{*}{R} & \multirow{2}{*}{\checkmark} \\
            \citep{9011028} & & --- & 5,336 & & No semantic labels & & \\
            \midrule
            \multirow{2}{*}{\textbf{DarkVision}} & Image & 900$\times$3 & 13,455 & 12 & 15-class boudning boxes & R & \checkmark\\
            & Video & 32$\times$3& 89,411 &  12 & 4-class boudning boxes & R & \checkmark\\
		\midrule
		\label{tab:statistics}
	\end{tabular}
	\vspace{-4mm}
\end{table*}

Faced with the urgent demands from both academia and industry, we build a two-path optical setup to capture low-light image/video and its bright reference simultaneously, and introduce a new dataset called DarkVision to supplement this field with a large-scale low-light image/video benchmark. This dataset contains images of 900 static scenes and videos from 32 dynamic scenes, captured at 5 illuminance levels, and offers pixel-wisely aligned bright reference to supervise image/video enhancement and object detection. With bright reference images, the DarkVision is manually annotated accurately, with 7,344 bounding boxes from 15 categories in the static subset and 320,667 bounding boxes from 4 object categories in the dynamic subset. Fig.~\ref{fig:ExampleImg} shows an exemplary static scene at 5 different illuminance levels and its bright reference, and Fig.~\ref{fig:ExampleVideo} is a video example. In Table~\ref{tab:statistics}, we compare our dataset and other low-light datasets, and show that DarkVision is unique in terms of larger data scale, more data types and richer annotations. In Section~\ref{sec:dataset}, we will present detailed characteristics of DarkVision.

Towards the goal to set up a benchmark, we comprehensively evaluate the performance of representative methods for two basic tasks---namely image/video enhancement and object detection---on the DarkVision through extensive experiments. Next, to demonstrate an exemplary application of DarkVision, we propose two simple yet effective strategies for the above two video-based tasks respectively. Concretely, we utilize the temporal and spatial consistency in video frames to assist in these video-based tasks. Experiment results demonstrate the challenges of DarkVision and the necessity of incorporating spatial and temporal priors simultaneously. We further investigate the relationship between high-level and low-level tasks based on our dual-task dataset and the coherent results verify that DarkVision is a reliable database for cross-task study with low-light data.


The main contributions of this paper are summarized as follows:
\begin{itemize}
	\item We build a portable two-path optical setup capable of capturing pixel-level registered bright-dark image pairs simultaneously and collect a large-scale low-light benchmark dataset composed of an image subset with 900 static scenes and a video counterpart with 32 dynamic scenes and 89,411 frames. All data are acquired at 5 illuminance levels by three different-grade cameras and have accurate annotations.
	\item We have evaluated representative methods on the DarkVision image and video subsets for both enhancement and object detection, constructing a benchmark to facilitate future research. Based on the extensive experiment results, we analyze the challenges of DarkVision and provide in-depth discussions on its advantages compared with other related datasets.
	\item To demonstrate an exemplary application of DarkVision, we propose to explicitly exploit the temporal consistency in low-light videos by designing simple yet effective strategies for enhancer and object detector respectively, which achieve better performance than frame-by-frame methods.
	\item We further investigate the relationship between high-level and low-level tasks and verify that our DarkVision can contribute to cross-task study with low-light data and the advancement of this challenging research area.
\end{itemize}

In the proceeding sections, we review the related work of low-light vision comprehensively in Section~\ref{sec:relatedwork}. Next, we describe the workflow of the database construction and its characteristics in Section~\ref{sec:dataset}. Afterwards, temporally consistent networks for video enhancement and object detection are quickly explained in Section~\ref{sec:multichannel} as a demonstration of DarkVision's application. In Section~\ref{sec:experiments} we conduct extensive experiments and analysis on the DarkVision benchmark, including the performance of baseline algorithms, our proposed strategies, mutual interplay between enhancement and object detection and real nighttime image enhancement results. Finally, in Section~\ref{sec:conclusion} we conclude the paper with a summary and discussions on the future work.

\section{Related Work}\label{sec:relatedwork}
\subsection{Low-light datasets}
Due to the rapid development of data-driven methods, high-quality and large-scale data are desperately needed. Recently, various efforts have been devoted into collecting low-light datasets targeting at image enhancement and semantic tasks.

A number of datasets are composed of low-light images collected from the Internet or existing public dataset and corresponding reference bright images created by further post-processing~\citep{LOH201930,shen2017msr,Wang_2019_CVPR,8259342}.  Shen et al.~\citep{shen2017msr} collected a dataset of 10,000 low-light and bright image pairs from UCID~\citep{10.1117/12.525375}, BSD~\citep{5557884} and Google search engine after careful selection and retouchment. Wang et al.~\citep{Wang_2019_CVPR} built a dataset (SICE) with 3,000 underexposed images (2,750 images for training and the rest for testing) collected from real scenes and the Internet. Each of these images is further processed using Adobe Lightroom so as to serve as its corresponding reference. Cai et al.~\citep{8259342} proposed a single image contrast enhancer (SICE) dataset that contains 4,413 low-contrast images of different exposures from 589 sequences and their corresponding high-quality reference images. These low-light images are clustered into different low-light conditions/levels but have no true bright reference. Recently, Lv et al.~\citep{lv2021attention} constructed a large-scale synthetic low-light dataset which contains 22,656 scenes with specially designed simulation strategies. Some other researchers capture low-light images/videos by changing exposure and ISO~\citep{Chen_2018_CVPR,anaya2018renoir,Chen2018Retinex,Jiang_2019_ICCV}. Chen et al. introduced See-in-the-Dark (SID) dataset~\citep{Chen_2018_CVPR} containing 5,094 short-exposure raw images and 424 long-exposure reference raw images with the illuminance being roughly between 0.2 lux and 5 lux. Similar to SID, RENOIR~\citep{anaya2018renoir} is a dataset consisting of low-light uncompressed natural images of 120 scenes with three cameras of different sensor sizes. LOw-Light (LOL) dataset~\citep{Chen2018Retinex} contains 500 low/normal-light image pairs captured in real scenes and 1000 synthetic ones for training.

In addition to these image datasets, efforts have also been paid in collecting low-light video datasets. Dark Raw Video (DRV) dataset~\citep{Chen_2019_ICCV} contains 202 static raw videos captured under environment with illumination ranging from 0.5 to 5 lux using the same approach as SID~\citep{Chen_2018_CVPR}. Wang et al.~\citep{9710730} adopted a mechatronic system to build a paired Seeing Dynamic Scene in the Dark (SDSD) dataset consisting of 150 scenes. However, such collection scheme is only applicable for scenes without moving objects. For capturing dark-bright pairs of dynamic scenes, Jiang et al. built the See-Moving-Objects-in-the-Dark (SMOID) dataset~\citep{Jiang_2019_ICCV} using a two-path optical system and captures 179 pairs of videos consisting of 35,800 extremely low-light raw images and their corresponding reference RGB counterparts. This acquisition approach is similar to ours, but the scale, quality, diversity, quantitative calibration are insufficient for a benchmark dataset. Besides, SMOID does not provide annotations, which limits the applications for high-level computer vision tasks.

Due to the difficulty of data acquisition, low-light datasets with rich annotations for high-level vision tasks (e.g., object detection and semantic segmentation) are rarely proposed. Exclusively Dark (ExDARK) dataset~\citep{LOH201930} includes 7,363 images collected from the Internet by searching low-light related keywords or retrieved from existing datasets like COCO~\citep{10.1007/978-3-319-10602-1_48}. These images contain 23,710 object instances of 12 classes with bounding box annotation. Yang et al.~\citep{9049390} proposed a low-light face detection dataset including 43,849 annotated faces as training and evaluation set. Following~\citep{Chen2018Retinex} and~\citep{9049390}, Liu et al.~\citep{liubenchmarking} collected a Vision Enhancement in the LOw-Light condition (VE-LOL) dataset serving both low/high-level vision with different under-exposure levels in real scenarios. The paired subset VE-LOL Low-Level Vision (VE-LOL-L) includes a total of 2,500 images with 1,000 synthesized from raw images in RAISE dataset~\citep{10.1145/2713168.2713194} while the unpaired subset VE-LOL High-Level Vision (VE-LOL-H) is composed of 10,940 under-exposure images with 91,330 annotated faces. To evaluate the performance of nighttime image semantic segmentation, Dai and Gool\citep{daytime:2:nighttime} and Sakaridis et al.\citep{9011028} prepared Nighttime Driving and Dark Zurich datasets with 50 and 151 labeled nighttime images respectively and large number of unlabeled samples.

Despite the above attempts on datasets collection that have contributed to the advancement in low-light image/video enhancement and understanding, there are still deficiencies with these datasets in four aspects, which limits further progress in low-light data analysis. (i) \textbf{Task diversity:} They are mainly oriented to either image enhancement task or single-class object detection. Other tasks such as video enhancement and object detection are seldom involved. (ii) \textbf{Illuminance-level calibration:} Most of existing datasets have ambiguous physical illuminance levels, if any, and mix all data together without discriminating illuminance levels. Since darkness levels contribute to different levels of challenges, it is difficult to conduct precise and quantitative evaluation regarding illuminance. (iii) \textbf{Scale:} Their scales are narrowly sufficient in terms of the number of images and scenes for enhancement task, but clearly deficient in the number of instance and category compared with recent datasets for object detection under normal illuminations. (iv) \textbf{Data type:} A majority of them only provide static images while perception of low-light dynamic video is equivalently or even more important. The challenge of constructing such a low-light image/video dataset for both enhancement and object detection mainly stems from the difficulty in capturing low-light and reference images simultaneously. To address this issue, we establish a two-path optical setup to build DarkVision for comprehensive low-light data analysis.

\subsection{Low-light image/video enhancement}

Low-light image/video enhancement is the most extensively explored topic in low-light data analysis. Here we review a series of methods of low-light data enhancement, which can be roughly divided into three groups based on intensity transformation, model prior, and end-to-end deep learning respectively.

\vspace{1mm}
\noindent\textbf{Intensity transformation based methods.~~~~} Low-light images are usually visually dark with extremely low contrast. Various intensity transformations can be applied for contrast improvement such as simple scaling and Gamma correction.
Histogram equalization (HE) seeks to stretch the dynamic range and enhance the contrast of an image by redistributing its histogram~\citep{HUMMEL1977184}. However, HE performs the histogram adjustment globally and has a tendency to cause unsatisfactory local contrast. To address this issue, adaptive histogram equalization (AHE) methods~\citep{PIZER1987355,pizer1990contrast,pisano1998contrast,reza2004realization} compute several histograms with each corresponding to a distinct region of the image, thus improving the local contrast and enhancing the edges in each region separately. Later, Kim \citep{580378} and Ibrahim and Kong \citep{4429280} proposed mean brightness preserving HE to overcome unnecessary visual deterioration. HE based on sub-image~\citep{754419} and sub-histogram~\citep{abdullah2007dynamic} have also been developed to preserve the original illuminance. In general, by paying more attention to local adaptivity, HE-based methods produce improved enhancement results. However, like other intensity transforms, most HE methods are of insufficient flexibility to local regions and might lead to under/over-exposure and amplified noise.

\vspace{1mm}
\noindent\textbf{Model based methods.~~~~}
Distinct from intensity-transformation based methods, model based enhancement methods introduce constraints from either a statistical or physical model.
(i)  {\em Statistical models} seek to fit the distribution of the pixel intensities in a latent sharp natural image. Celik and Tjahjadi~\citep{5773086} proposed to use inter-pixel contextual information and to find a histogram mapping that covers large intensity range. Li et al. ~\citep{8451278} proposed a three-component generalized Gaussian mixture model to fit the histogram of the illuminance image. Lee et al.~\citep{6615961} achieved contrast enhancement by seeking a layered difference representation of 2D histograms. Differently, Liang et al.~\citep{7352346} proposed a novel  method to estimate the illumination by iteratively solving a nonlinear diffusion equation.
Another line of research is to standardize the Poisson noise which is predominant in low-light scenarios by Variance Stabilizing Transform (VST)~\citep{7491301}, implement Gaussian denoiser~\citep{dabov2007image} and apply inverse transformation.
(ii) {\em Physical models} establish the relation between low-light images/videos with latent clean images/videos. Retinex theory~\citep{land1971lightness} assumes an image can be decomposed into two components--- reflectance and illumination, and the latter is estimated and removed to enhance the image/video~\citep{jobson1997properties,597272}.
To reduce unnatural illumination and over-enhancing artifacts, many improvements have been made over traditional Retinex model~\citep{6512558,FU201682,Fu_2016_CVPR,guo2016lime}.
The Retinex model is then refined by additionally estimating a noise map \citep{8304597} and introducing noise suppression~\citep{8351427}.
Another physical model is absorption light scattering model (ALSM)~\citep{8737871}. Alternatively by exploiting the statistical properties in natural videos, Ko et al. ~\citep{ko2017artifact-free} proposed a temporal similarity and guide map based algorithm.
In general, model based methods achieve impressive results in contrast enhancement and noise suppression. However, as the models and the related priors are hand-crafted or over-simplify the complicated illumination conditions and noise characteristics in the real world, the adaptability to large variety of data is unsatisfactory.

\vspace{2mm}
\noindent\textbf{Deep-learning based methods.~~~~}
Deep-learning based methods have become the mainstream due to their outstanding performance and efficient inference. Most works focus on enhancing a single low-light image to a visually pleasant result by designing elaborate network archetectures. Lore et al.~\citep{LORE2017650} proposed a deep auto-encoder based network to adaptively brighten images. Some other works~\citep{8305143,shen2017msr,8478390,8373911} utilized multiscale features to assist in low-light image enhancement. Further, Lv et al.~\citep{lv2018mbllen} proposed to extract rich features up to different levels and apply enhancement via multiple subnets and finally produce the output image via multi-branch fusion. To address raw measurements, Chen et al.~\citep{Chen_2018_CVPR} and Jiang et al.~\citep{Jiang_2019_ICCV} proposed end-to-end convolutional networks to enhance the data directly without demosaicing. Retinex module has also been fused into deep networks to maintain good signal structure ~\citep{shen2017msr,Chen2018Retinex,10.1145/3343031.3350983,10.1145/3343031.3350926}, and adversarial training was applied in the design of low-light enhancement networks~\citep{9334429,8803328} to suppress visual artifacts. Another direction explores camera noise model and generates large-scale paired simulated data for training~\citep{Wang_2019_ICCV,Wei_2020_CVPR}. Other approaches include deep hybrid network~\citep{8692732}, deep recursive band network (DRBN)~\citep{Yang_2020_CVPR}, attention-guided network~\citep{lv2021attention}, convolutional long short-term memory (ConvLSTM) network~\citep{xiang2019an}, frequency-based decomposition and-enhancement model~\citep{Xu_2020_CVPR} and zero-reference deep curve estimation (Zero-DCE)~\citep{Guo_2020_CVPR} etc. Directly applying single image enhancement methods to low-light videos may suffer from color distortion and flickering artifacts, so researchers introduce cascading network~\citep{9156652}, kernel prediction networks (KPN)~\citep{Mildenhall_2018_CVPR,9156965}, recurrent networks~\citep{10.1007/978-3-030-01267-0_33,Wang_2019_ICCV}, self-consistency loss~\citep{Chen_2019_ICCV}, and 3D convolution~\citep{lv2018mbllen,Jiang_2019_ICCV} to burst denoise or enhance videos by considering temporal information explicitly.



\begin{figure*}[t]
	\centering
	\begin{minipage}[ht]{0.29\textwidth}
		\subfigure[]{
			\begin{tabular}{c}
				\includegraphics[width=0.98\linewidth]{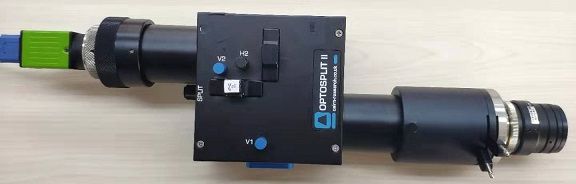} \\
				\includegraphics[width=0.98\linewidth]{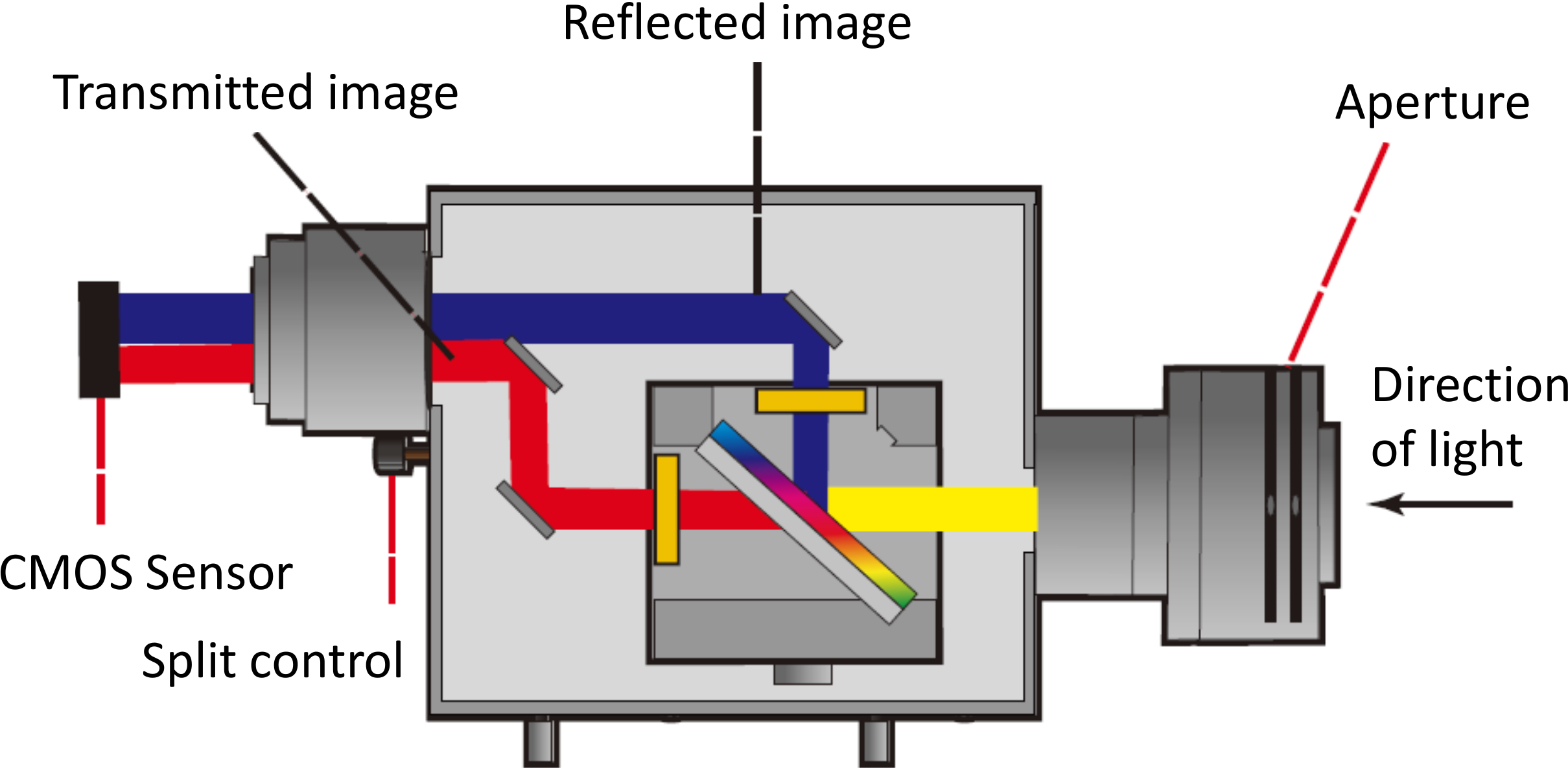}
			\end{tabular}
		}
	\end{minipage}
	\hspace{5mm}
	\begin{minipage}[ht]{0.64\textwidth}
		\subfigure[]{
			\includegraphics[width=0.98\linewidth]{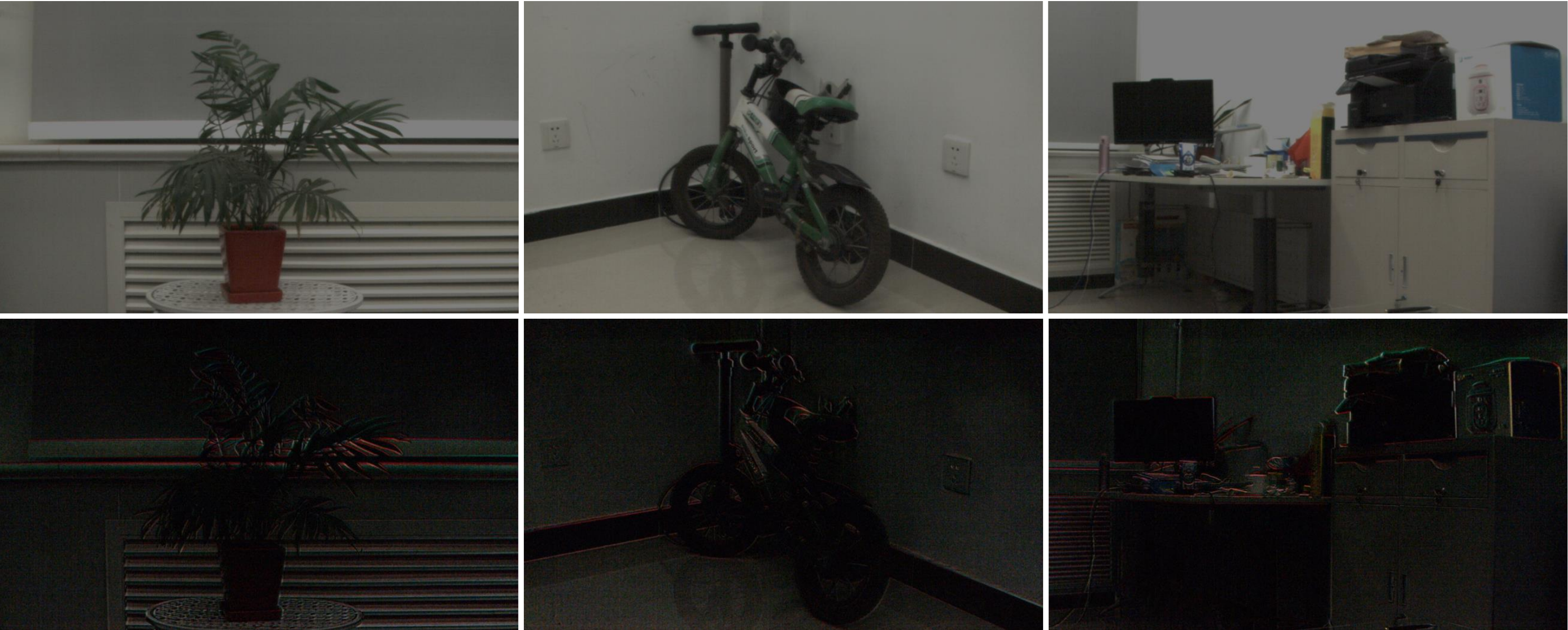}
		}
	\end{minipage}
	\caption{The imaging setup for database constrcution and registration between the low-light image and its bright counterpart in a snapshot. (a) The setup we built for capturing bright-dark pair of images (top) and its optical light path (bottom). The image sensor in our setup receives a latent image through the split optics where half travels through an ND filter. Therefore, the resolution of the images is 1/2 size in width. (b) Bright reference images (top) and the residue between two channels after photometric calibration (bottom). These images are captured by the low-end camera (JAI GO-5100C-USB).}
	\label{fig:setup}
\end{figure*}

\subsection{High-level tasks and interplay with image enhancement under low-light environment}
Vision tasks at differnt levels are not independent of each other. Recently, high-level vision tasks and its assistance for enhancement have been studied with related datasets and methodology. In the following part, we review current progress in low-light semantic understanding.

Object detection in low-light scenarios is important, but collecting an annotated low-light dataset is quite challenging. To address this issue, Sasagawa and Nagahara~\citep{10.1007/978-3-030-58589-1_21} proposed a domain adaptation method for low-light object detection alternatively. Researchers made efforts to study the relation between high-level tasks and image enhancement. One direction is to introduce guidance from high-level tasks into image/video enhancement. Chen et al.~\citep{5636218} proposed a video enhancement method by classifying each pixel as foreground or background and extracting moving objects, which are then enhanced with bilateral filter and Retinex algorithm. Similarly, Hao et al.~\citep{8996493} used saliency theories to enhance the specific region of interest (ROI) by fusing generated saliency map and the enhanced image. Some other researchers attempted to study tasks at different levels jointly. For example, Liu et al.~\citep{8960640} explored the interaction between image denoising and high-level vision tasks by extracting contextual information at different scales and cascading two modules for image denoising and various high-level tasks respectively to benefit both tasks. Considering that face is an important pattern in low-light vision, Yang et al.~\citep{9049390} analyzed the effect of enhancement for low-light face detection and compared the results of different algorithms. Liu et al.~\citep{liubenchmarking} jointly conducted low-light enhancement and face detection by passing the features extracted by the enhancement module to the successive layer with the same resolution of the detection module. Overall, these methods offer valuable insights into the mutual benefits between high-level and low-level vision tasks, but further studies are still largely confined by available datasets.

\section{The DarkVision Database}\label{sec:dataset}

\subsection{Acquisition setup}
\noindent\textbf{Optical design.~~~~}
To capture paired low-light and reference images for both static and dynamic scenes, we develop a setup by inserting a light splitter (OptoSplit II, Andor Corp Industries) between the lens and camera sensor to record the paired data simultaneously within a snapshot. As illustrated in Fig.~\ref{fig:setup}, the device splits the entering light into two paths with equal power and maps them to two separate regions of the pixel array. Optical filters such as attenuation filter or band-pass filter can be inserted into two slots of the splitter for respective path. For our equipment, we insert a neutral density (ND) filter which attenuates light evenly into one slot while leave the other empty, to capture an image with low and high illumination parts respectively. ND filters with different transmission rates (TRs) are adopted to vary the attenuation level, ranging at 3.2\%, 1.6\%, 0.8\%, 0.4\%, and 0.2\%. We apply these filters for each scene sequentially so that we obtain images of the same scene with five darkness levels. Exemplar images of a static scene are shown in Fig.~\ref{fig:ExampleImg} and a dynamic scene in  Fig.~\ref{fig:ExampleVideo}.
Lens is mounted to the splitter via standard C-mount, and we change different lenses for varying target scenes. The simultaneous capture mode can avoid camera motion and possible environment variations between low-light image and its reference.

To adapt to more natural scenes, we use three lenses with focal length being 12.5mm, 25mm and 50mm to capture objects of different sizes (large, medium and small). In order to enrich data diversity, we use three different-grade cameras from different manufacturers to capture each scene. They are JAI GO-5100C-USB  ($2464\times2056$ pixels with 3.45 \textmu m pixel size), FLIR BFS-U3-50S5C ($2448\times2048$ pixels with 3.45 \textmu m pixel size) and PCO panda 4.2 C  ($2048 \times 2048$ pixels with 6.5 \textmu m pixel size). The first two are based on CMOS sensor while the last one is an sCMOS camera, which is of high quantum efficiency (QE), low readout noise and thus widely applicable for photon limited applications. All three cameras output 12-bit color images at 30 fps for both image and video acquisition. The exposure time is kept as 33ms during acquisition. The aperture size varies depending on the light conditions. For example, at noon when illumination is strong, the aperture size varies from f/8 to f/16, while at dusk when light is dim, the aperture size is set from f/1.4 to f/5.6. The illuminance of the scenes in the dataset approximately ranges from 100 lux to 2,000 lux. Due to the fact that under extremely low-light circumstances, noise and color distortion also severely degrade reference image quality, we apply no gain or Gamma correction to record the raw data so as to offer more choices for further processing.


\begin{figure}[t]
	\centering
	\begin{minipage}[ht]{0.98\textwidth}
		\subfigure[]{
			\includegraphics[width=0.49\textwidth]{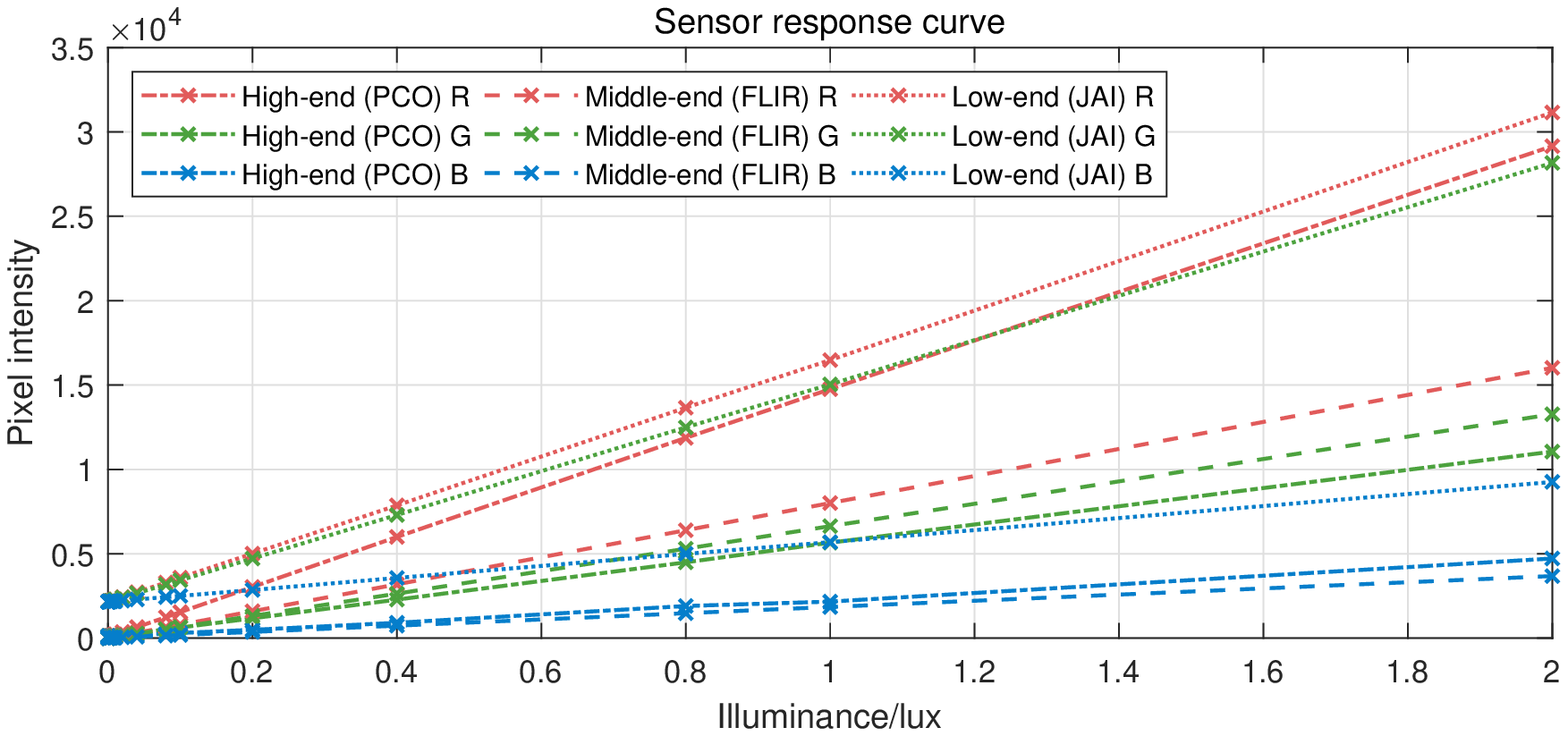}
		}\\
		\subfigure[]{
            \includegraphics[width=0.24\textwidth]{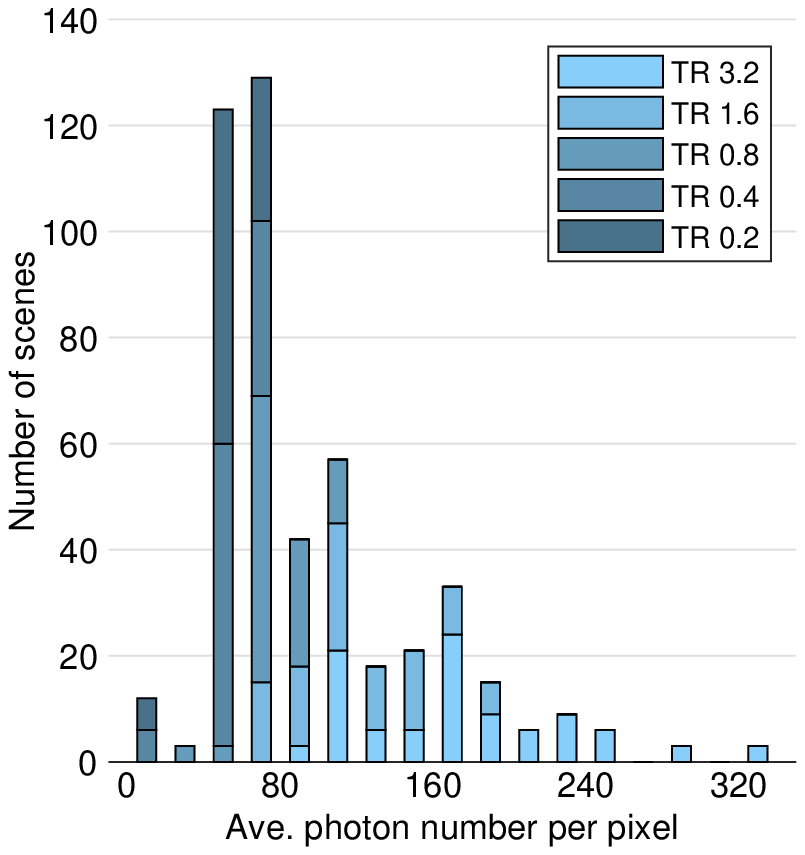}
            }
            \subfigure[]{
            \includegraphics[width=0.24\textwidth]{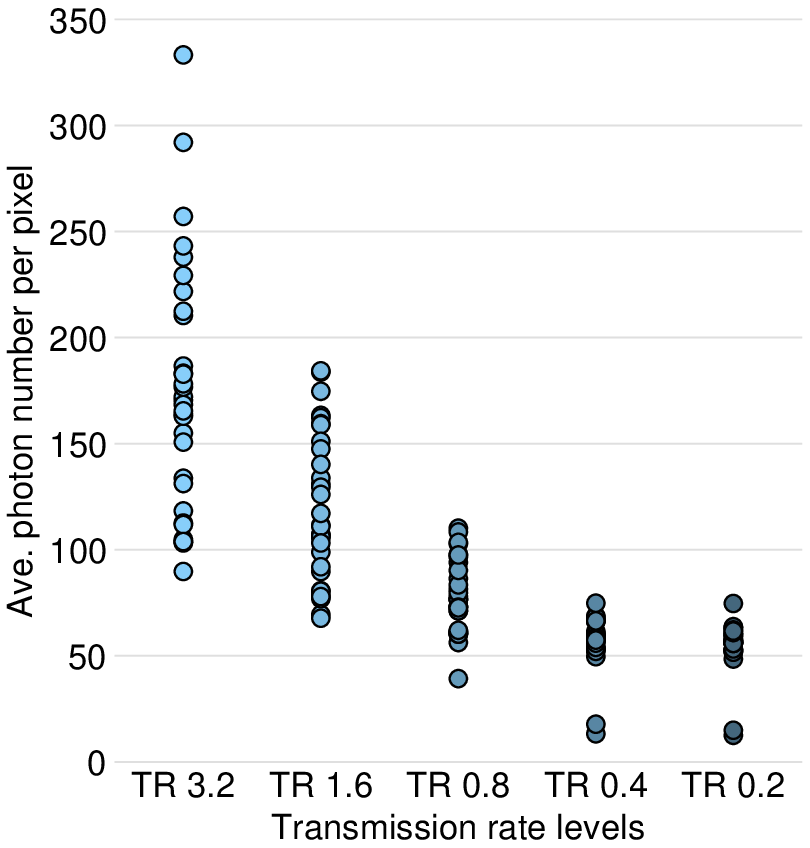}
		}
	\end{minipage}
	\caption{{(a) The response curves of the three used camera sensors, describing the mapping between the radiance arriving at the sensor and the output intensity. Conversely, one can infer the illuminance at each scene point from the corresponding intensity of the recorded image. (b) and (c) Photon number distribution of all the scenes in DarkVision database, calculated via the intensity value and response curves. Each dot in (c) denotes a scene.}}
	\label{fig:calibration}
\end{figure}

\vspace{1mm}
\noindent\textbf{Sensor response calibration.~~~~}
In order to provide a quantitative analysis on the illumination level, we calibrate image sensor response curve to the radiation arriving at the pixel entries, by which we can infer the received photons statistically from the image intensity. Specifically, we placed sensors at a distance away from an illuminance-adjustable large flat light source with uniform known Lambertion radiance. By varying the illumination from 0 to 2 lux, we acquire 20 corresponding readout images at each illuminance, and then calculate the average pixel intensity and plot the illuminance-readout relation of each sensor, as shown in Fig.~\ref{fig:calibration}. We calculate these curves via raw data with no gain or Gamma correction applied. The lowest calibrated illuminance is 0.00001 lux. However, the camera enters the 'dead zone' when illuminance drops to 0.001 lux, i.e., the curve becomes flat and the intensity is close to that of 0 lux (pure noise).
With the response curve, one can infer the illuminance of the scene from the input image accordingly, after subtracting the 'dark image' captured at 0 lux. Such quantitative evaluation is important for testing the algorithm performances at different illuminance levels. We provide the photon number distribution of all scenes which is calculated via the sensor response curve. For most scenes, the average photon number falls into the range of 20-240. Low-light data with such low photon number have never been captured and calibrated to the best of our knowledge. We hope our DarkVision benchmark can push the research boundary of low-light imaging to extremely dark scenarios.

\begin{figure*}[t]
	\centering
	\includegraphics[width=0.96\textwidth]{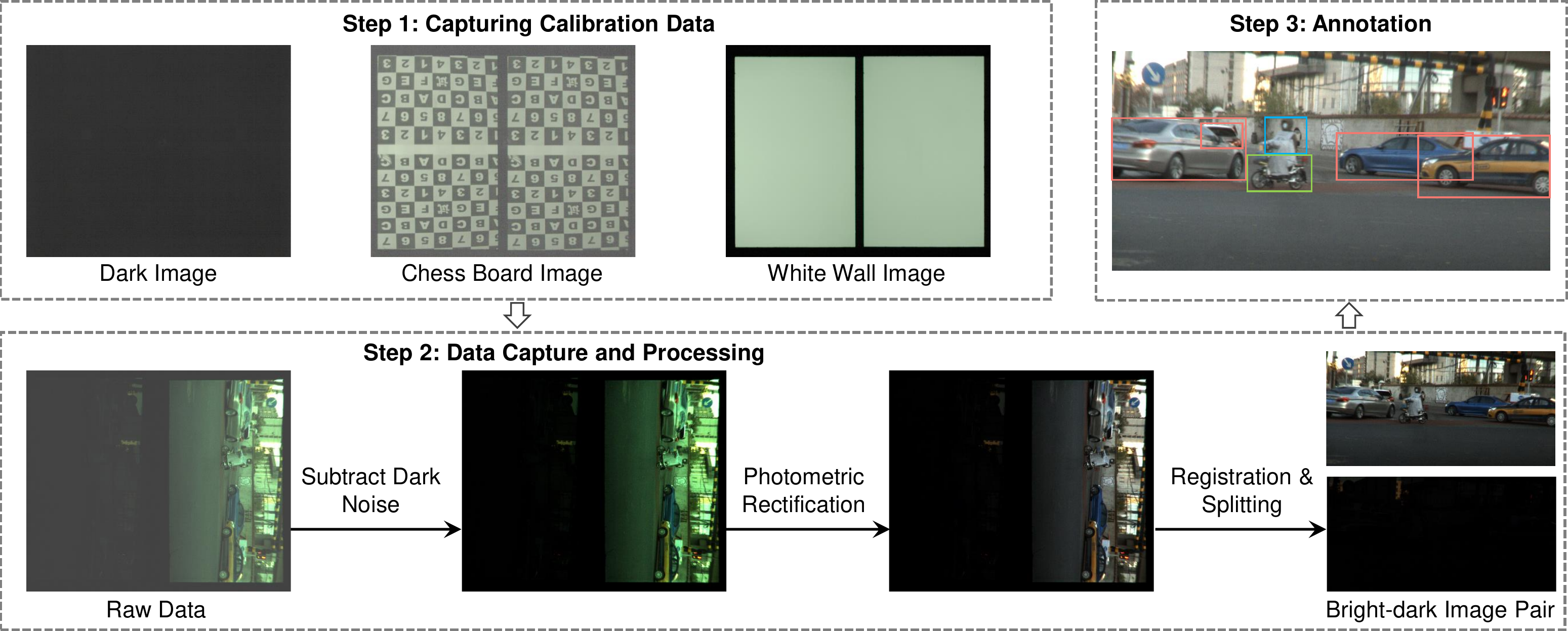}
	\caption{The pipeline for capture, post-processing and annotation of the DarkVision benchmark. Here we apply scaling to the images in Steps 1 and 2 and additional Gamma transformation to the image in Step 3 for better visualization.}
	\label{fig:pipeline}
\end{figure*}

\begin{figure}[htbp]
	\centering
	\vspace{2mm}
	\includegraphics[width=\linewidth]{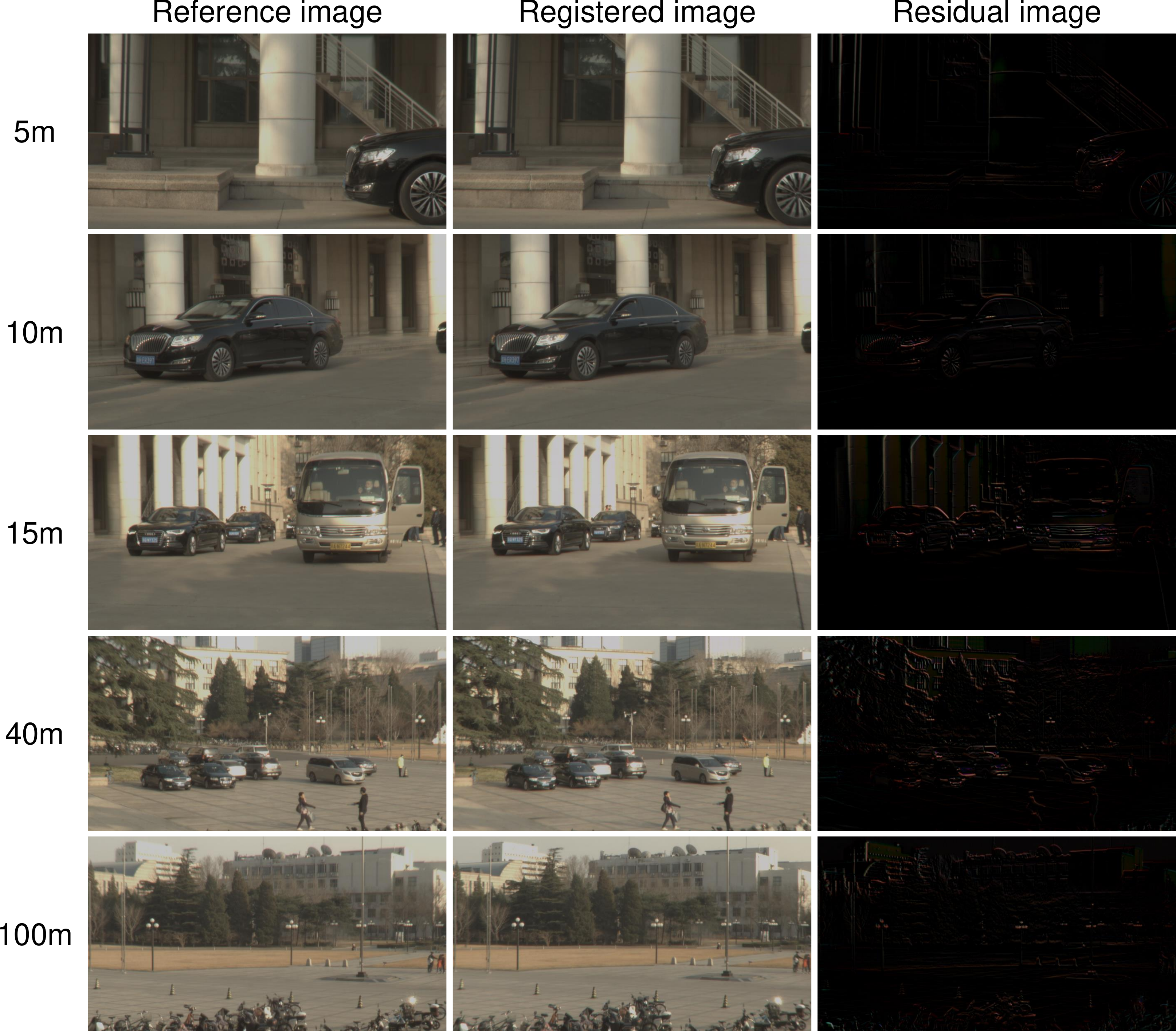}
	\caption{{Validation of registration between two channels. We set one sub-image as reference (left column) and warp another for registration (middle column), with the residue demonstrating registration error (right column). These exemplar images are captured by the low-end camera.}}
	\label{fig:residule}
\end{figure}

\subsection{Pre-processing and annotation}
As aforementioned, each captured image contains two parts: low-light part and its reference counterpart. To produce a pair of pixel-wise aligned images, we apply a series of data pre-processing steps to raw sensor outputs, as shown in Step 1 and Step 2 in Fig.~\ref{fig:pipeline}.

\vspace{1mm}
\noindent\textbf{Calibration / rectification.~~~~} Firstly, a dark image is captured beforehand and subtracted from the captured data to compensate the black level, which is negligible for bright image but would severely deteriorate the quality of low-light images. Specifically, we keep camera settings fixed and acquire 100 black level images with lens cap on and then take average over these images to get the desired dark image. Secondly, we rectify the photometric difference between low-light and reference regions. As there exists slight photometric difference in brightness between the two channels of our device, which is caused by the imperfections of the beam splitter and non-uniform response at different regions of the sensor, we use a Lambertian white board to compensate this between-channel difference. We capture the image of a Lambertian white board, and remove the photometric difference across the sensor by calculating the ratio image with respect to this ``white image".

\vspace{1mm}
\noindent\textbf{Registration.~~~~} Afterwards, we perform registration and cropping to extract the image pair. We capture a chessboard pattern image without inserting any ND filter beforehand, feed it for similarity based MATLAB registration algorithm and then crop low-light and reference image pair of proper size.  The details are as follows. First, we split the full latent image into two halves and relay them to two non-overlapping regions on the sensor, which contains left and right subimages and surrounding black background. Next, the two halves are used to calculate a transform matrix either by automatic similarity-based registration or manually selecting correspondence points. Specifically, we set the left halve fixed, warp the right halve using the calculated transform matrix and crop the contents from the black edges. When performing registration on captured images, we apply the same transform matrix on the reference subimages so that noise characteristics in low-light images remain unchanged. The comparison between residual image from two paths (no attenuation filter applied) shown in Fig.~\ref{fig:setup}(b) demonstrates that the geometric alignment is of high precision and photometric rectification dramatically reduces the difference between two paths. Moreover, we performed an experiment to demonstrate the accuracy of our registration algorithm. We specially captured scenes with varying depths (ranging from 5m to 100 m) and register the sub-image pairs using a scene-independent registration matrix calculated from a chessboard snapshot, with the registration accuracy demonstrated in Fig.~\ref{fig:residule}. The average intensity of residual image is around $3\%-5\%$ of the reference image's intensity, which is sufficiently small to be neglected for pixel-wise metrics such as PSNR. These results again validate the feasibility of our image acquisition and processing pipeline.

\vspace{1mm}
\noindent\textbf{Annotation.~~~~}
Our database provides reliable bright counterpart as reference, which can be used for annotation right after pre-processing as displayed in Step 3 of Fig.~\ref{fig:pipeline}. Similar to many other object detection datasets, bounding-box labels are annotated by a well trained professional team following a strict guideline. The annotations of image and video subsets are different. For static scenes, the images of different illuminance levels are aligned and annotation is performed only once. Differently, the videos at 
varying illuminance levels are  captured at the same location but in a sequential manner, 
so we annotate all the sequences.

\subsection{Data collection and statistics}
\begin{figure}[t]
	\centering
	\begin{minipage}[ht]{0.98\textwidth}
		\subfigure[]{
			\includegraphics[width=0.255\textwidth]{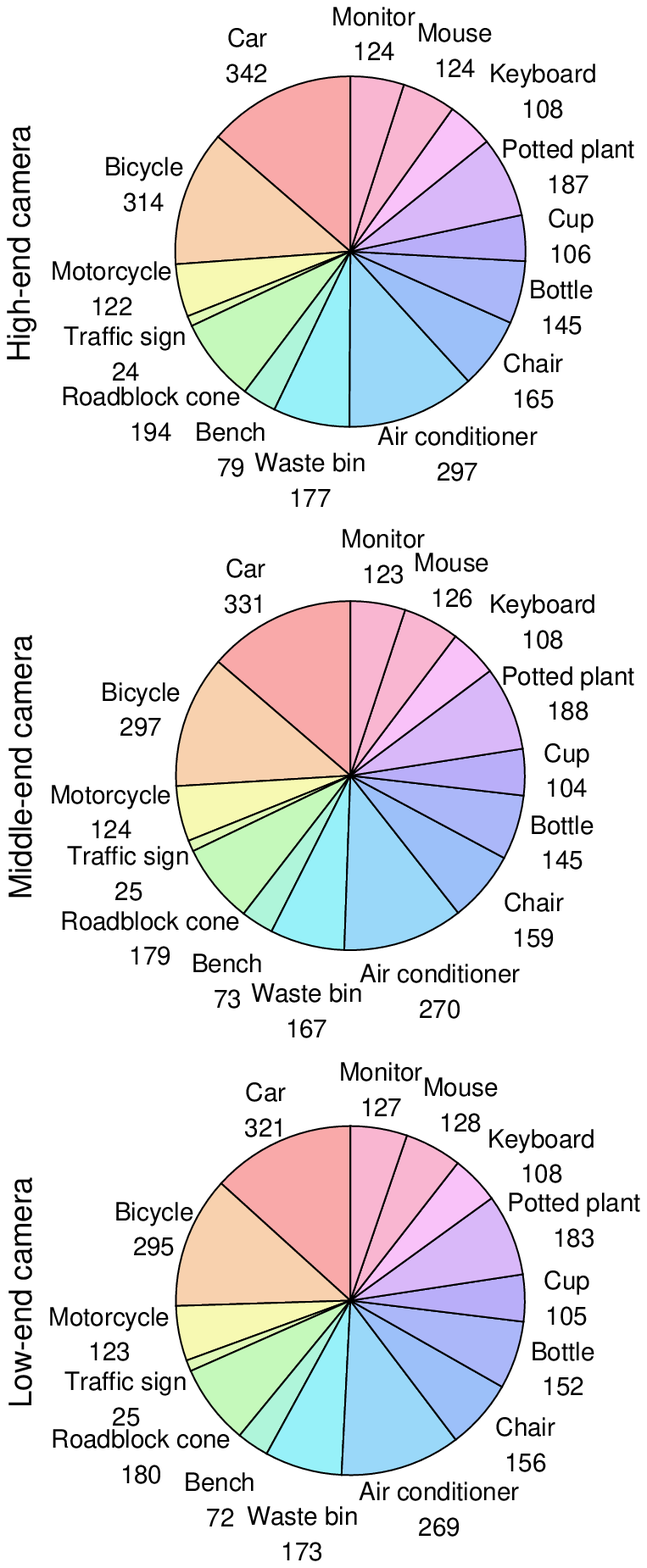}
		}
		\subfigure[]{
			\includegraphics[width=0.23\textwidth]{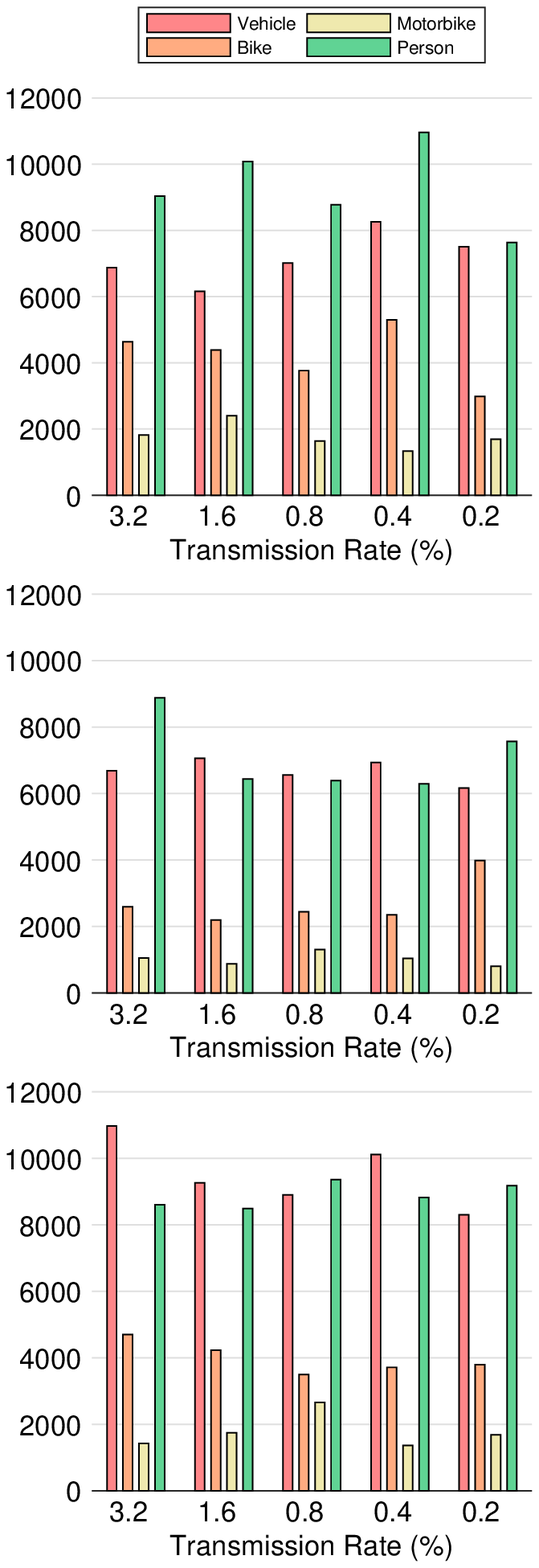}
		}
	\end{minipage}
	\caption{The number of object instances of 15 different categories in our static image subset (a), and 4 categories in the video subset (b). Note that the static images at different illuminance levels are aligned, while there exist slight among-camera differences in the static image subset and among-illumination-level differences in the video subset.}
	\label{fig:class-num}
\end{figure}

\begin{figure*}[t]
	\centering
	\vspace{-1mm}
	\includegraphics[width=0.96\textwidth]{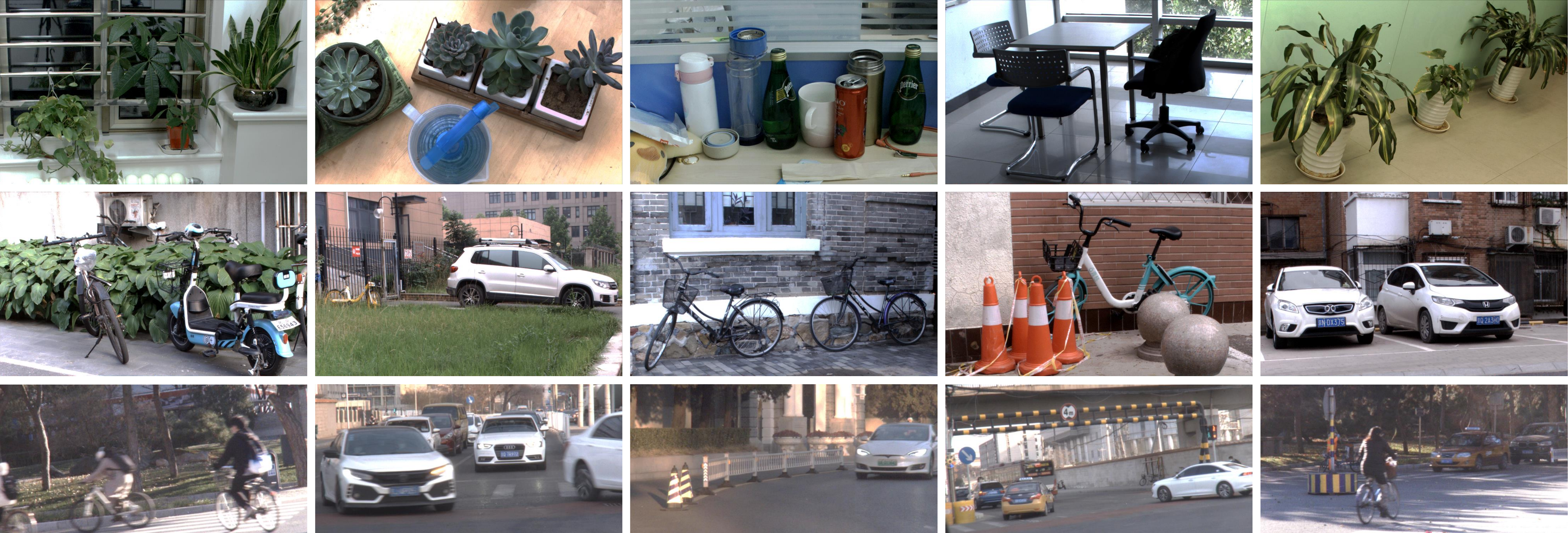}
	\caption{Sample scenes from our DarKVision dataset. Images in the three rows are from static image subset (indoor), static subset (outdoor) and dynamic video subset. We apply linear scaling and Gamma transformation for better visualization.}
	\label{fig:showcase}
	\vspace{-2mm}
\end{figure*}

Considering the data availability, we mainly collected the images/videos on campus, in residential areas and on roads. The DarkVision dataset is composed of two parts---images (static scenes) and videos (dynamic scenes). The image subset includes 900 images from 15 categories of objects (e.g. person, car, and cup), with the instance number of each category listed in Fig.~\ref{fig:class-num} (a). These object categories are carefully selected via jointly considering the benchmark diversity, category density together with the wide availability in daily life. For the video subset, we capture 32 video clips (89,411 frames in total) of 4 categories with good availability---person, car, bike and motorbike---at 30 fps. The statistics of the instance number in the video benchmark are displayed in Fig.~\ref{fig:class-num} (b). We showcase sample scenes from our dataset in Fig.~\ref{fig:showcase}. All data are collected at 5 different transmission levels for each camera. The images/frames are recorded in 12-bit TIFF format (less quantization error than 8-bit data) and raw data is also provided to avoid signal loss by compression, which is especially important for low-light data. It is worth noting that we captured low-light and reference pairs at one specific transmission level at a time. Therefore, for dynamic scenes the videos at different levels are acquired successively and thus there exist slight differences in the number of object instances. Given the large instance number, we argue that the differences are negligible for illuminance-related analysis.
\begin{table}[t]
    \centering
    \caption{Number of images and instances for train/test sets used in our experiment.}
    \vspace{1mm}
    \label{tab:set}
    \small
    \begin{tabular}{lllll}
        \midrule
        \multirow{2}{*}{Set} &\multicolumn{2}{c}{Image subset}& \multicolumn{2}{c}{Video subset}\\ \cline{2-5}
        &$\#$ Images&$\#$ Instances&$\#$ Frames&$\#$ Instances\\
        \toprule
        {Train} & 10,740 & 29,410 & 69,910 & 246,143 \\
        {Test}  & 2,715 & 7,310 & 19,501 & 74,524  \\
        {Total}	& 13,455 & 36,720 & 89,411 & 320,667 \\
        \midrule
    \end{tabular}
\end{table}

We partition the image subset into 716 and 184 scenes (images) and the video subset into 26 scenes (around 4,600 frames) and 6 scenes (around 1,300 frames) for training and testing respectively for each illuminance level and for each camera. Test set has totally different scenes from the train set to prevent well-trained networks from “memorizing” similar scenes and leading to inflated high performance. Table~\ref{tab:set} displays the number of images and instances of all the data in terms of dataset partition.


\section{An Exemplary Application of DarkVision for Temporally Consistent Enhancer \& Detector}\label{sec:multichannel}
Existing image enhancement and object detection algorithms can be directly applied to our DarkVision dataset. However, their performance under such extremely low-light conditions cannot be guaranteed. Considering the temporal correlation that may provide valuable prior under severe noise degeneration, here we demonstrate an exemplary application of DarkVision to validate its advantages as a large dark video dataset with reliable ground truth. Specifically, we design a multi-frame video enhancer and a detector that explicitly incorporate temporal consistency in low-light videos. The results in Section \ref{sec:experiments} validate that although simple, the two methods outperform single-frame methods. Besides, they show comparative or even better performance than some multi-frame methods in certain cases. More importantly, temporal information should be specially considered for low-light video analysis. Next, we will present the details of the proposed methods.



\subsection{Temporally consistent dark video enhancer (TC-DVE)}
\begin{figure}[ht]
	\centering
	\vspace{-1mm}
	\includegraphics[width=0.9\linewidth]{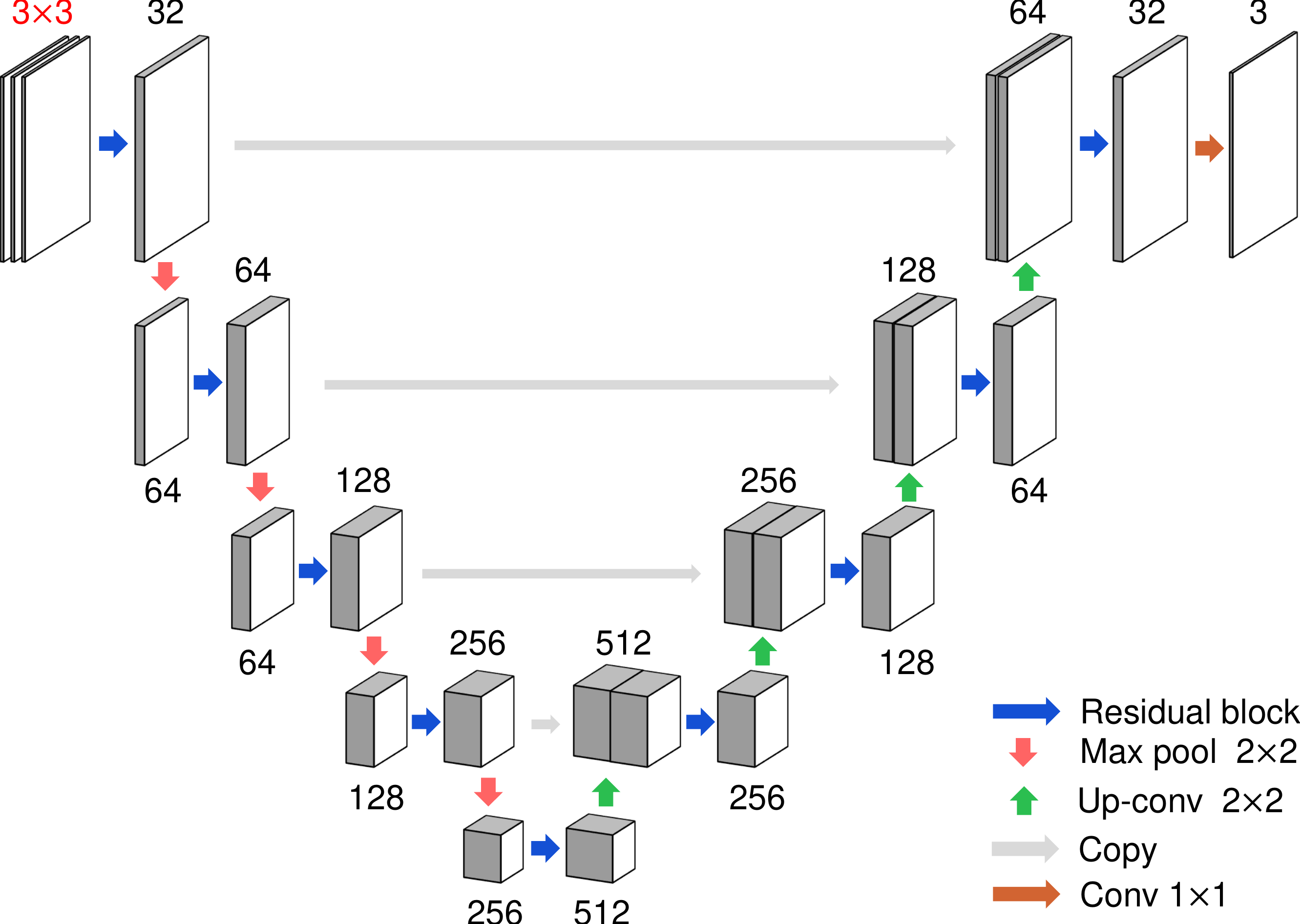}
	\vspace{-1mm}
	\caption{The network structure of the proposed temporally consistent enhancer, which exploits temporal correlation by using multi-channel input and residual blocks.}
	\label{fig:unet}
	\vspace{-2mm}
\end{figure}

Recent research works have exploited convolutional networks to enhance the quality of low-light images and achieved impressive performance. However, single-frame based enhancers may suffer from flickering and color inconsistency when applied to low-light videos in a frame-by-frame manner. To overcome these drawbacks, we propose a temporally consistent dark video enhancer with the network architecture illustrated in Fig.~\ref{fig:unet}. We adopt UNet~\citep{10.1007/978-3-319-24574-4_28} as the basic backbone for our image restoration task so that we can utilize the encoder-decoder structure for feature extraction and noise suppression. Further, we integrate residual blocks~\citep{He_2016_CVPR} for better representation and training. To leverage temporal cues in neighboring frames, we pile up a number of adjacent frames as input and encourage the network to output the restored result of the middle frame. In our implementation, we concatenate three neighboring frames (indexed by \textit{i}-1, \textit{i} and \textit{i}+1) along the channel dimension to reconstruct the middle frame, i.e., \textit{i}th one. This scheme exploits bidirectional temporal information related to the target frame.

\begin{figure*}[t]
	\centering
	\includegraphics[width=0.96\textwidth]{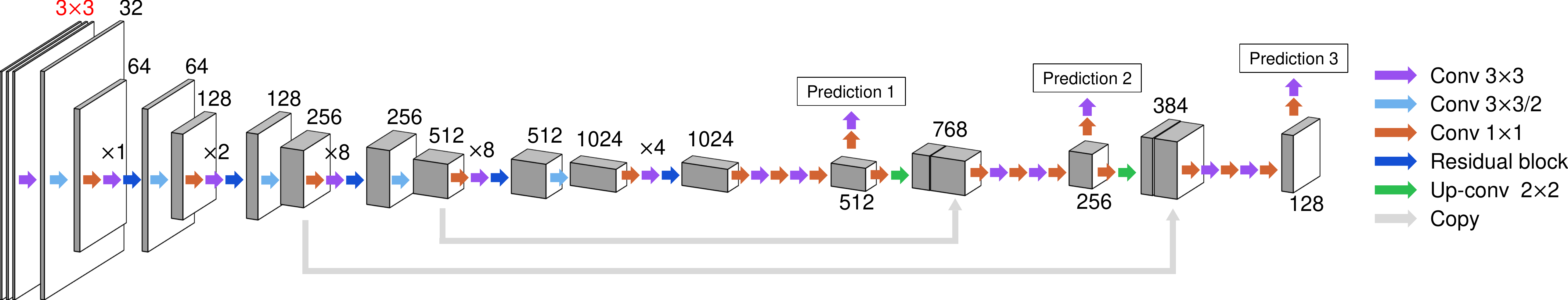}
	\vspace{-1mm}
	\caption{We take the network structure of TC-YOLO as an example (due to its simplicity compared with the other two detectors) to show the proposed temporally consistent object detection approach, which exploits temporal information effectively and is a general architecture compatible with any existing image-based object detectors.}
	\label{fig:mcYOLO}
	\vspace{-1mm}
\end{figure*}

\subsection{Temporally consistent dark video detector (TC-DVD)}

Extending single-image detectors to videos in a frame-by-frame manner is a common and intuitive practice with off-the-shelf detectors in hand (e.g., Faster R-CNN, webcam demos). However, the accuracy of detection suffers from motion blur, video defocus, rare poses etc. Since nearby frames in a video share resemblance in object category, object position and background, a number of related research works exploit this temporal consistency to design detectors specially for videos and have achieved superior performance to frame-by-frame scheme. For example, Kang et al. \citep{7780464} adopted tubelet-based CNN for object detection from videos. Zhu et al.~\citep{zhu2017deep} proposed deep feature flow (DFF) framework that utilized a flow field to propagate feature maps from multiple frames. Later, they further improved per-frame features by aggregation of nearby features along the motion paths by introducing flow-guided feature aggregation (FGFA)~\citep{8237314}. Different from previous works that emphasize more on the temporally nearby frames, Wu et al.~\citep{wu2019sequence} devised a novel sequence level semantics aggregation (SELSA) module to aggregate features in the full-sequence level.
Chen et al.~\citep{Chen_2020_CVPR} proposed a memory enhanced mechanism that aggregates global and local frames to assist in the middle frame detection. Recently, Gong et al.~\citep{gong2021temporal} proposed a novel Temporal ROI Align operator to extract features from other frames feature maps for current frame proposals by utilizing feature similarity.

Alternatively, like TC-DVE, we propose a simple yet effective strategy to adapt single-image object detectors for dark videos. The strategy can be incorporated into almost all state-of-the-art single image detectors. Similar to TC-DVE, we concatenate a couple of adjacent frames along the channel dimension as input to the network which outputs the object category(s) and position(s) of the middle frame. Without loss of generality, we use three representative image detectors---YOLO, RetinaNet, and Faster R-CNN---as examples and name them as TC-YOLO, TC-RetinaNet and TC-Faster R-CNN. Taking YOLO as an example, the strategy is illustrated in Fig.~\ref{fig:mcYOLO}. In our implementation, we empirically pile up $3$ color frames in total, i.e. 9 channels for input. We modify the kernel size of the first layer convolution of the backbone to match our strategy. In Section \ref{sec:experiments}, we perform rigorous experiments to show the effectiveness of our method on DarkVision benchmark.

\section{Experiments}\label{sec:experiments}

In this section, we perform extensive experiments on three different tasks for low-light vision: (1) object detection, (2) image/video restoration, (3) mutual interplay between object detection and restoration. For the first two tasks, we quantitatively evaluate the performance of state-of-the-art methods and provide a benchmark on our DarkVision dataset. We also test the performance of our temporally consistent strategy for both tasks on dark videos. For the last task, we propose preliminary approaches conducting object detection and restoration jointly and report the results on DarkVision, intending to study the potential mutual benefits between two tasks in dark scenarios and provide insights into this challenging problem. Lastly, to validate the generalization of the models trained from DarkVision on real low-light images, we present the enhancement result of a real nighttime image captured by a smartphone. The following part describes experiment implementation details, results and analysis.

\subsection{Evaluation for object detection}

Image/video object detection is an extensively studied topic and a large number of deep networks have been proposed and achieved excellent performance in recent years. For the image subset, we evaluate the most representative detectors: one-stage detectors---RetinaNet~\citep{fu2019retinamask} and YOLO~\citep{redmon2018yolov3} and two-stage detector---Faster R-CNN~\citep{ren2015faster}.
For the video counterpart, we additionally evaluate 5 multi-frame detectors---DFF~\citep{zhu2017deep}, FGFA~\citep{8237314}, SELSA~\citep{wu2019sequence}, MEGA~\citep{Chen_2020_CVPR} and Temporal ROI Align~\citep{gong2021temporal} and our proposed temporally consistent dark video detectors (i.e., TC-Faster R-CNN, TC-RetinaNet and TC-YOLO).

To provide a comprehensive analysis, we conduct training/testing on all the data groups---3 cameras, 5 different illuminance levels, both static images and videos---separately. In terms of quantitative performance evaluation, we use mean average precision (mAP) from the Pascal VOC evaluation toolkit~\citep{EveringhamEGWWZ15} as the metric. All experiments are conducted using PyTorch and Python on NVIDIA GeForce RTX 3090 GPUs. We follow the default data augmentation in these networks, including random resizing, flipping and cropping etc.


\begin{table*}[t!]
    \centering
    \caption{The quantitative evaluation (mAP) of the baseline object detection algorithms on the static data benchmark. {\bf{Bold}} indicates the best score.}
    \begin{center}
    \resizebox{1\textwidth}{!}{
    \begin{tabular}{l|lllll|lllll|lllll}
        \midrule
        & \multicolumn{5}{c|}{High-end camera} & \multicolumn{5}{c|}{Middle-end camera} & \multicolumn{5}{c}{Low-end camera}\\
        TR (\%) & 3.2 & 1.6 & 0.8 & 0.4 & 0.2 & 3.2 & 1.6 & 0.8 & 0.4 & 0.2 & 3.2 & 1.6 & 0.8 & 0.4 & 0.2 \\
        \toprule
        Faster R-CNN & 58.5 & 56.9 & 42.4 & 29.7 & 31.4 & 47.1 & 34.5 & 20.2 & 9.4 & 10.1 & 35.7 & 30.2 & 19.1 & 12.7 & 10.2 \\
        RetinaNet  & 57.8 & 54.2 & 39.9 & 30.0 & 30.3 & 41.3 & 36.0 & 19.7 & 8.1 & 9.3 & 40.4 & 32.9 & 21.0 & 11.1 & 9.1\\
        YOLO  & \bf{60.9} & \bf{60.7} & \bf{52.8} & \bf{45.0} & \bf{36.5} & \bf{55.0} & \bf{53.1} & \bf{44.6} & \bf{29.6} & \bf{30.3} & \bf{50.1} & \bf{49.6} & \bf{34.6} & \bf{28.9} & \bf{29.6}\\
        \midrule
    \end{tabular}}
    \end{center}
    \label{tab:det_results_static}
\end{table*}

\begin{table*}[t!]
    \centering
    \caption{The quantitative evaluation (mAP) of the baseline object detection algorithms on the dynamic data benchmark. {\bf{Bold}} indicates the best score.}
    \vspace{-3mm}
    \begin{center}
    \resizebox{1\textwidth}{!}{
    \begin{tabular}{l|lllll|lllll|lllll}
        \midrule
        & \multicolumn{5}{c|}{High-end camera} & \multicolumn{5}{c|}{Middle-end camera} & \multicolumn{5}{c}{Low-end camera}\\
        TR (\%) & 3.2 & 1.6 & 0.8 & 0.4 & 0.2 & 3.2 & 1.6 & 0.8 & 0.4 & 0.2 & 3.2 & 1.6 & 0.8 & 0.4 & 0.2 \\
        \toprule
        Faster R-CNN & 55.7 & 48.9 & 36.3 & 28.3 & 13.8 & 54.1 & 41.9 & 29.3 & 9.6 & 4.8 & 43.5 & 34.8 & 26.6 & 7.9 & 5.2\\
        RetinaNet & 55.0 & 48.1 & 32.3 & 22.1 & 12.8 & 51.0 & 43.5 & 28.7 & 11.9 & 8.6 & 39.9 & 35.5 & 26.1 & 10.3 & 3.9\\
        YOLO & 40.1 & 32.8 & 19.0 & 13.2 & 8.2 & 29.5 & 29.0 & 23.3 & 8.9 & 5.2 & 34.4 & 25.5 & 20.2 & 9.5 & 9.8\\
        \midrule
        TC-Faster R-CNN & 57.6 & 49.3 & 37.1 & \bf{33.1} & 16.2 & 59.4 & 43.9 & 36.0 & 10.1 & 5.8 & 43.6 & \bf{39.3} & 27.1 & 8.9 & 13.0\\
        TC-RetinaNet & 56.9 & 49.6 & \bf{39.4} & 25.8 & 16.1 & 52.1 & 44.3 & 43.5 & \bf{14.6} & 9.6 & 40.7 & 37.9 & \bf{31.5} & 10.4 & 7.4\\
        TC-YOLO & 43.5 & 39.3 & 26.4 & 20.1 & 12.6 & 38.8 & 37.5 & 26.1 & 12.1 & 7.6 & 35.1 & 32.1 & 26.8 & 14.2 & \bf{17.7}\\
        \midrule
        DFF & 49.3 & 45.8 & 18.9 & 18.7 & 3.7 & 53.4 & 38.8 & 16.9 & 3.8 & 4.3 & 40.0 & 29.8 & 22.6 & 8.1 & 7.4\\
        FGFA & 59.6 & 52.2 & 30.0 & 28.8 & 17.0 & \bf{71.9} & 48.3 & 30.8 & 8.8 & 8.1 & \bf{51.4} & 35.1 & 23.4 & 16.9 & 8.8\\
        SELSA & 63.9 & \bf{55.3} & 38.9 & 31.7 & \bf{19.9} & 68.1 & \bf{56.3} & 42.7 & 13.4 & 11.2 & 48.5 & 35.7 & 26.7 & \bf{19.8} & 16.5\\
        MEGA & 52.5 & 40.2 & 27.6 & 26.6 & 13.5 & 52.5 & 51.4 & 39.1 & 11.8 & 10.5 & 30.5 & 30.1 & 13.9 & 13.4 & 11.2\\
        Temporal ROI Align& \bf{64.4} & 53.6 & 37.7 & 30.4 & 18.6 & 70.2 & 52.3 & \bf{45.9} & 13.9 & \bf{11.3} & 51.1 & 37.8 & 29.0 & 19.1 & 16.3\\
        \midrule
    \end{tabular}}
    \end{center}
    \label{tab:det_results_dynamic}
\end{table*}


\begin{figure*}[t]
	\centering
	\includegraphics[width=0.98\textwidth]{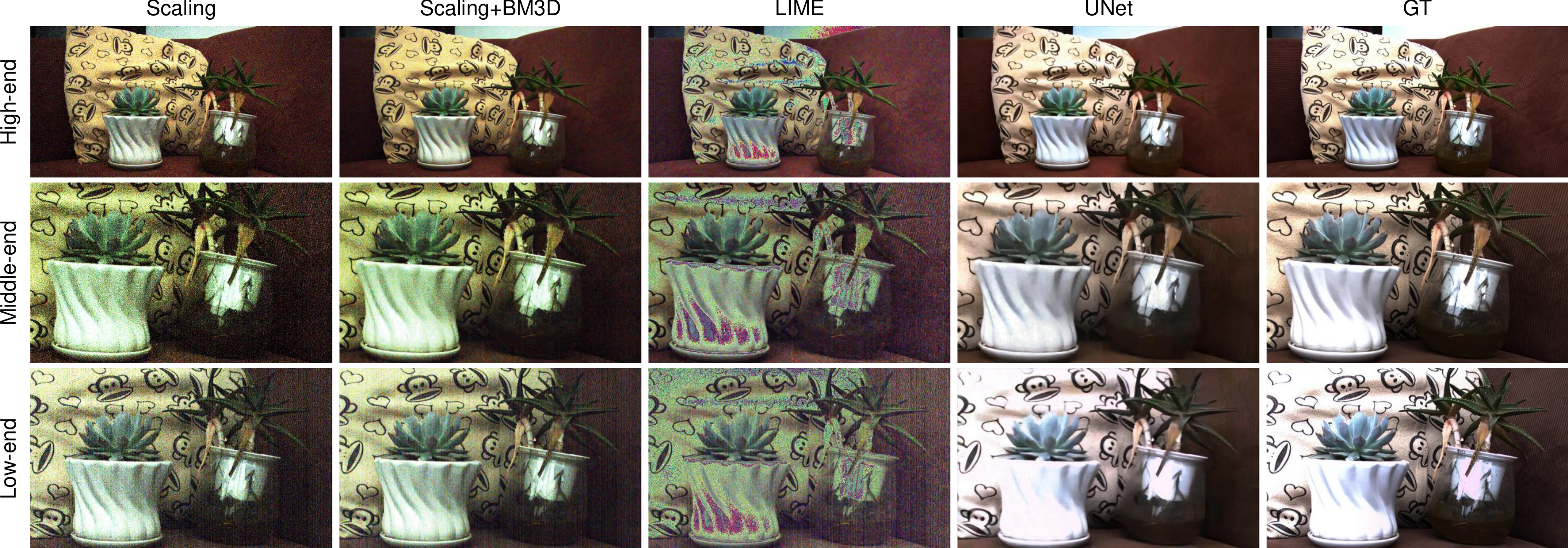}
	\caption{The restoration results and visual performance comparison of different methods on exemplar static images.}
	\label{fig:result_img_rec}
\end{figure*}

\begin{figure*}[t]
	\centering
	\vspace{-2mm}
	\includegraphics[width=0.98\textwidth]{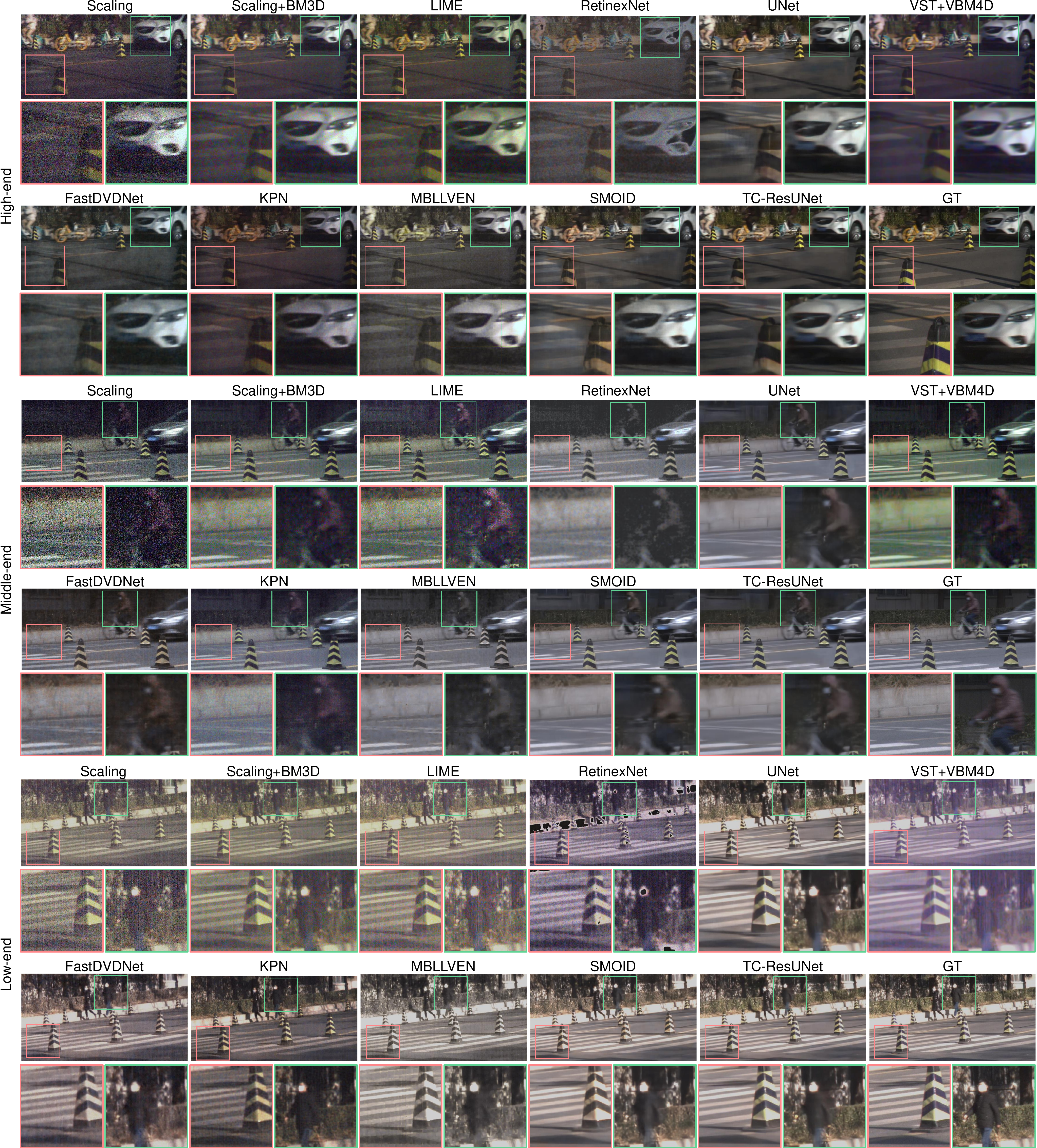}
	\caption{The restoration results and visual performance comparison of different methods on exemplar video frames.}
	\label{fig:result_vid_rec}
\end{figure*}


\begin{figure*}[t]
	\centering
	\includegraphics[width=0.98\textwidth]{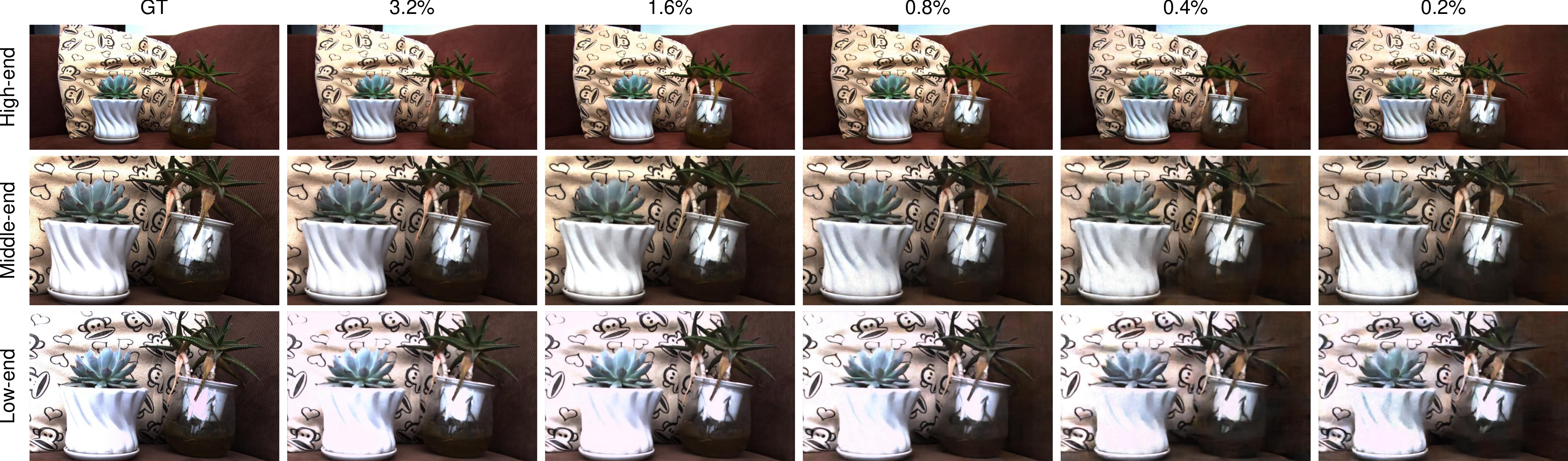}
	\caption{The restoration results and visual performance comparison of different illuminances on an exemplar static image. The restoration results are produced by UNet.}
	\label{fig:result_rec_comp_STAR}
	\vspace{-2mm}
\end{figure*}

\begin{figure*}[t]
	\centering
	\includegraphics[width=0.98\textwidth]{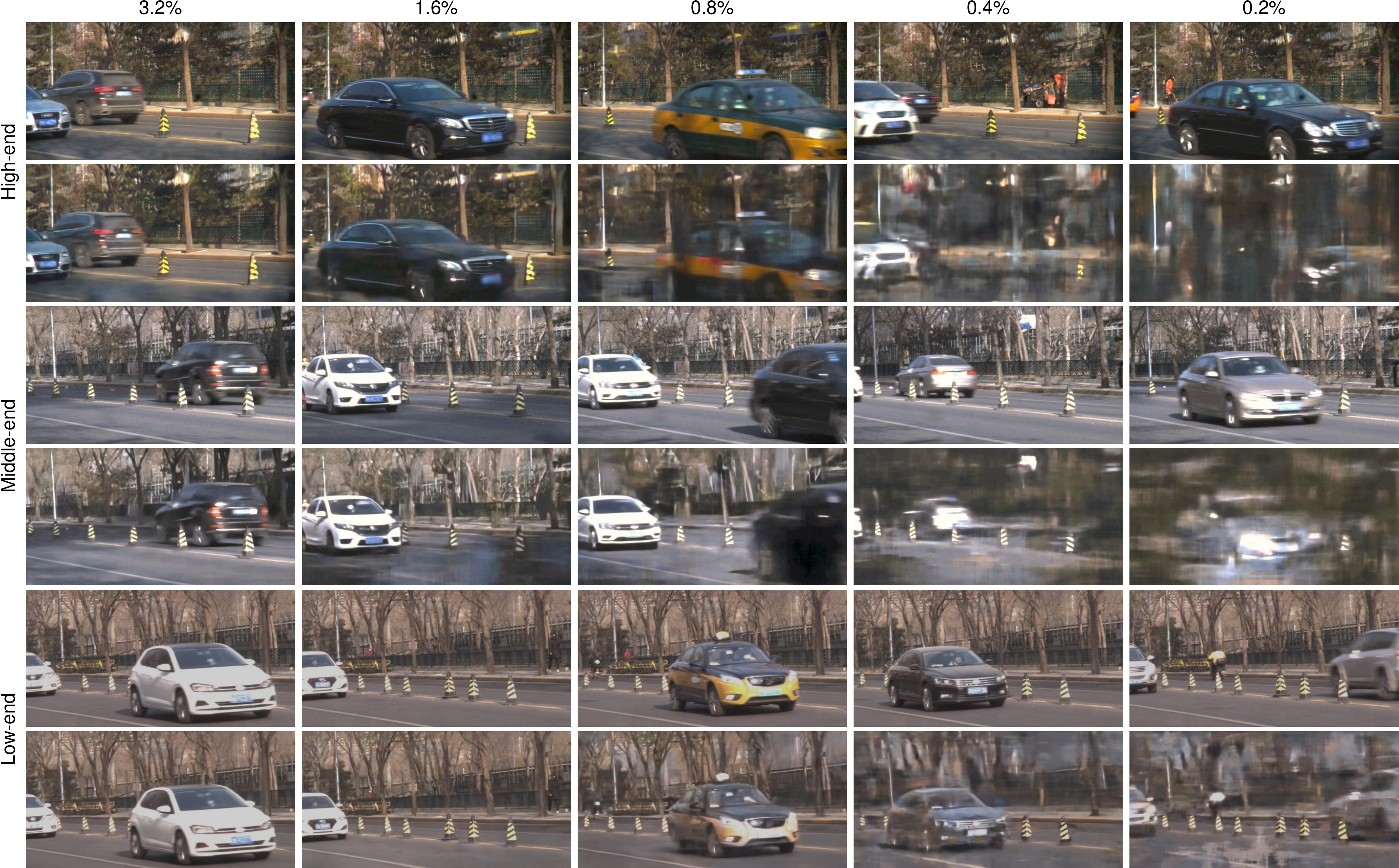}
	\caption{The restoration results and visual performance comparison of different illuminances on an exemplar video frame. For each camera, upper row is GT and lower row restoration results. The restoration results are produced by TC-ResUNet.}
	\label{fig:result_rec_comp_STAR_vid}
	\vspace{-2mm}
\end{figure*}

\vspace{1mm}
\noindent\textbf{Quantitative analysis.~~~~}
Tables \ref{tab:det_results_static} and \ref{tab:det_results_dynamic} present the experiment results for image subset and video subset respectively.
As expected, the detection performance degenerates with the decrease of TR in general.
For example, with Faster R-CNN implemented on high-end camera data we can find that: the mAP (\%) drops from 58.5 to 31.5 on the image subset when TR changes from the highest to lowest; there is also a steady drop between the video detection results on different illuminance levels (55.7, 48.9, 36.3, 28.3, 13.8 correspondingly), and the highest mAP is 41.9 points higher than the lowest, which is a huge difference. Results of different detectors and different cameras display a similar trend.
This reveals our DarkVision provides different-level challenges for image/video object detection, which is beyond all the existing datasets.

Among three cameras, performance trend is roughly consistent with their image quality, especially under low TRs. When we use the same set of attenuation rates, almost all detectors achieve best results on high-end camera data, second best on middle-end camera data and worst on low-end camera data. This implies the necessity of including different cameras for building low-light dataset and the importance of camera selection according to the application and budget.


By comparing the detection performance before and after introducing temporal consistency constraint (i.e., YOLO vs. TC-YOLO, RetinaNet vs. TC-RetinaNet, and Faster R-CNN vs. TC-Faster R-CNN, on low-light videos), we observe a universal mAP improvement under all settings---varying TRs and cameras. Especially, at low TR and with low camera sensitivity, this improvement is more obvious. For example, the performance of Faster R-CNN on the low-end camera data, at TR 3.2\% there is a 0.1 point rise in mAP while at TR 0.2\%, the mAP improvement is 7.8. This demonstrates the greater benefits from multi-frame strategy in more severely degraded low-light video detection. Since the low-light data suffers from extremely low SNR, single frame hardly contains sufficient information for accurate object detection. When considering the temporal consistency existing in neighboring frames, much information gain is achieved in this challenging scenario, which leads to better detection performance. This observation indicates that effectively using both spatial and temporal correlation by well-designed network architectures could be a promising way for perception in low-light videos.

On the video subset, we also evaluate a number of well-designed multi-frame detectors. Overall, no detector can achieve consistently best result under different settings: previously proposed video detectors perform well under higher and lower TRs, but degenerate in the middle brightness; in comparison to these normal-light video detectors that were elaborately devised, our simple TC-DVDs still performs best in 6 of 15 settings. These figures reveal the challenges of DarkVision and designing low-light video detectors that can adapt to different illuminations and cameras. 


Moreover, we observe that the baseline detectors exhibit different performance on image and video subsets. Specifically, on the single-image subset, YOLO achieves the highest mAP while on video subset leads to unsatisfactory performance. The good performance of YOLO on the static subset can be attributed to the following causes. Firstly, YOLO divides the input image into grids and conducts prediction within each cell, which often leads to degenerated detection precision of small-size or crowed objects. This feature is favourable for our image dataset which contains more indoor scenes with larger objects and similar appearances. Secondly, due to simple network structure, YOLO may be easier to train on static subset which contains limited number of training samples compared to Faster R-CNN and RetinaNet with complex network designs. On the video subset with richer training data, Faster R-CNN and RetinaNet result in better performance than YOLO.

\subsection{Evaluation for image/video restoration}
For low-light image restoration, we evaluate the performance of four algorithms, including simply scaling, BM3D~\citep{dabov2007image}, LIME~\citep{guo2016lime} and deep-learning based UNet from SID~\citep{Chen_2018_CVPR}. For the video subset, we additionally evaluate 7 algorithms which are RetinexNet~\citep{Chen2018Retinex}, VST+VBM4D~\citep{10.1117/12.872569}, FastDVDNet~\citep{9156652}, KPN~\citep{Mildenhall_2018_CVPR}, MBLLVEN~\citep{lv2018mbllen}, SMOID~\citep{Jiang_2019_ICCV} and the proposed temporally consistent enhancement network (TC-ResUNet). The 7 methods are video enhancement ones except RetinexNet. It is worth noting that we conduct the following adjustments to optimize the performance of these algorithms that are not specially designed for such extremely dark images/videos. 
For BM3D which is designed mainly for denoising images captured under normal illumination, we scale the brightness of low-light images to match that of the bright reference before noise suppression. For LIME, we adopt the default setting provided in the source code and apply BM3D and scaling after enhancement. When training RetinexNet, we use the linearly scaled ground truth as supervision signal and apply Gamma transformation to the network's output. For VST+VBM4D, we apply VST to standardize the Poisson noise before implementing VBM4D to denoise these video frames, and afterwards apply inverse transformation to return the denoised frames to their original range. When implementing FastDVDNet, we scale the brightness to transform low-light videos to 8-bit so as to adapt to the network. When training KPN, MBLLVEN and SMOID, we follow the default settings. When training the end-to-end UNet and our proposed TC-ResUNet, we use Adam optimizer to train the model with the initial learning rate of 0.0001. The training process lasts for 800 epochs, and the learning rate is scaled down by 0.1 at the 400th epoch. For deep-learning-based methods, we randomly crop $256 \times 256$ patches from full-resolution data (no resizing implemented) and perform random flipping for data augmentation. As for quantitative evaluation of restoration results, we adopt widely used Peak Signal-to-Noise Ratio (PSNR) and Structural SIMilarity (SSIM)\citep{wang2004image} as metrics. Before calculating these metrics of video subset, we apply white balance, linearly scaling and Gamma transformation to adapt the restored frames for visual observation, and these post-processing steps are implemented for all the video restoration results as well. 


\begin{table*}[t!]
    \centering
    \caption{The quantitative evaluation of the baseline image restoration algorithms in terms of PSNR (dB) and SSIM on the static data benchmark. {\bf{Bold}} indicates the best score.}
    \vspace{-3mm}
    \begin{center}
    \resizebox{1\textwidth}{!}{
    \begin{tabular}{l|rrrrr|rrrrr|rrrrr}
        \midrule
        \bf{PSNR (dB)} & \multicolumn{5}{c|}{High-end camera} & \multicolumn{5}{c|}{Middle-end camera} & \multicolumn{5}{c}{Low-end camera}\\
        TR (\%) & 3.2 & 1.6 & 0.8 & 0.4 & 0.2 & 3.2 & 1.6 & 0.8 & 0.4 & 0.2 & 3.2 & 1.6 & 0.8 & 0.4 & 0.2 \\
        \toprule
        Scaling & 28.58 & 26.32 & 21.09 & 18.27 & 18.34 & 26.95 & 23.89 & 18.49 & 16.11 & 16.16 & 24.09 & 21.00 & 15.72 & 13.74 & 13.73\\
        Scaling+BM3D & 29.57 & 27.82 & 22.89 & 19.66 & 19.73 & 29.44 & 26.60 & 20.23 & 17.06 & 17.13 & 24.72 & 21.70 & 16.18 & 14.07 & 14.06\\
        LIME & 27.47 & 26.17 & 23.13 & 21.56 & 21.62 & 27.66 & 25.84 & 22.87 & 21.81 & 21.82 & 22.84 & 20.96 & 17.84 & 16.81 & 16.83\\
        UNet & \bf{32.41} & \bf{32.13} & \bf{31.17} & \bf{30.14} & \bf{30.30} & \bf{34.11} & \bf{33.50} & \bf{32.69} & \bf{31.17} & \bf{31.43} & \bf{29.56} & \bf{29.35} & \bf{28.24} & \bf{26.74} & \bf{26.66}\\
        \midrule
        \bf{SSIM} & \multicolumn{5}{c|}{High-end camera} & \multicolumn{5}{c|}{Middle-end camera} & \multicolumn{5}{c}{Low-end camera}\\
        TR (\%) & 3.2 & 1.6 & 0.8 & 0.4 & 0.2 & 3.2 & 1.6 & 0.8 & 0.4 & 0.2 & 3.2 & 1.6 & 0.8 & 0.4 & 0.2 \\
        \toprule
        Scaling & 0.71 & 0.54 & 0.22 & 0.11 & 0.12 & 0.52 & 0.34 & 0.11 & 0.06 & 0.06 & 0.41 & 0.22 & 0.07 & 0.04 & 0.04\\
        Scaling+BM3D & 0.78 & 0.64 & 0.31 & 0.16 & 0.17 & 0.69 & 0.53 & 0.20 & 0.09 & 0.09 & 0.45 & 0.25 & 0.07 & 0.04 & 0.04\\
        LIME & 0.69 & 0.56 & 0.35 & 0.28 & 0.28 & 0.64 & 0.51 & 0.34 & 0.30 & 0.30 & 0.41 & 0.27 & 0.15 & 0.12 & 0.12\\
        UNet& \bf{0.90} & \bf{0.90} & \bf{0.88} & \bf{0.86} & \bf{0.86} & \bf{0.89} & \bf{0.88} & \bf{0.86} & \bf{0.83} & \bf{0.84} & \bf{0.85} & \bf{0.84} & \bf{0.82} & \bf{0.78} & \bf{0.78}\\
         
        \midrule
    \end{tabular}}
    \end{center}
    \label{tab:enh_results_static}
\end{table*}



\begin{table*}[t!]
    \centering
    \caption{The quantitative evaluation of the baseline image restoration algorithms in terms of PSNR (dB) and SSIM on the dynamic data benchmark. {\bf{Bold}} indicates the best score.}
    \vspace{-3mm}
    \begin{center}
    \resizebox{1\textwidth}{!}{
    \begin{tabular}{l|rrrrr|rrrrr|rrrrr}
        \midrule
        \bf{PSNR (dB)} & \multicolumn{5}{c|}{High-end camera} & \multicolumn{5}{c|}{Middle-end camera} & \multicolumn{5}{c}{Low-end camera}\\
        TR (\%) & 3.2 & 1.6 & 0.8 & 0.4 & 0.2 & 3.2 & 1.6 & 0.8 & 0.4 & 0.2 & 3.2 & 1.6 & 0.8 & 0.4 & 0.2 \\
        \toprule
        Scaling & 19.91 & 18.37 & 16.10 & 13.40 & 12.53 & 14.45 & 12.03 & 9.30 & 7.68 & 7.44 & 17.42 & 14.34 & 11.90 & 9.89 & 10.11\\
        Scaling+BM3D & 20.66 & 19.52 & 17.83 & 15.28 & 14.38 & 16.27 & 14.22 & 11.74 & 9.81 & 9.15 & 16.94 & 14.04 & 11.83 & 9.98 & 10.25\\
        LIME & 19.26 & 17.95 & 16.21 & 14.28 & 13.86 & 15.54 & 13.66 & 11.80 & 10.65 & 10.28 & 16.47 & 14.22 & 12.57 & 10.95 & 11.04\\
        RetinexNet & 21.84 & 18.23 & 21.45 & 20.34 & 18.88 & 21.65 & 21.13 & 21.28 & 18.06 & 16.68 & 17.95 & 16.96 & 16.59 & 16.61 & 15.34\\
        UNet & 25.49 & 24.82 & 24.30 & 21.79 & 20.80 & 27.38 & 25.45 & 22.98 & 18.51 & 17.78 & 26.83 & 24.09 & \bf{21.64} & 19.39 & 18.21\\
        VST+VBM4D & 22.02 & 21.93 & 20.61 & 17.96 & 16.82 & 23.88 & 22.04 & 17.96 & 14.53 & 13.60 & 20.16 & 18.63 & 15.33 & 13.97 & 13.55\\
        FastDVDNet & 22.92 & 23.08 & 22.27 & 19.54 & 19.32 & 26.56 & 25.68 & 22.16 & \bf{19.76} & 17.54 & 23.37 & 19.52 & 14.36 & 15.34 & 13.81\\
        KPN & 22.65 & 20.16 & 15.64 & 13.85 & 13.14 & 27.58 & \bf{25.95} & 10.75 & 10.56 & 10.04 & 25.23 & 15.04 & 10.09 & 8.52 & 8.47\\
        MBLLVEN & 22.20 & 22.20 & 21.46 & 19.67 & 14.10 & 24.43 & 23.99 & 20.71 & 11.89 & 8.44 & 24.28 & 22.46 & 19.82 & 14.60 & 15.61\\
        SMOID & 25.39 & 24.96 & 24.20 & \bf{22.09} & 21.04 & 27.12 & 25.86 & \bf{23.74} & 19.73 & 17.92 & 26.32 & 23.77 & 20.85 & 17.85 & 17.77 \\
        TC-ResUNet & \bf{25.92} & \bf{25.02} & \bf{24.49} & 21.99 & \bf{21.21} & \bf{27.70} & 25.78 & 23.13 & 19.03 & \bf{18.28} & \bf{27.39} & \bf{24.46} & 21.51 & \bf{19.52} & \bf{18.35} \\
        \midrule
        \bf{SSIM} & \multicolumn{5}{c|}{High-end camera} & \multicolumn{5}{c|}{Middle-end camera} & \multicolumn{5}{c}{Low-end camera}\\
        TR (\%) & 3.2 & 1.6 & 0.8 & 0.4 & 0.2 & 3.2 & 1.6 & 0.8 & 0.4 & 0.2 & 3.2 & 1.6 & 0.8 & 0.4 & 0.2 \\
        \toprule
        Scaling & 0.30 & 0.20 & 0.11 & 0.06 & 0.05 & 0.11 & 0.05 & 0.03 & 0.01 & 0.01 & 0.20 & 0.12 & 0.06 & 0.03 & 0.03\\
        Scaling+BM3D & 0.51 & 0.40 & 0.27 & 0.17 & 0.14 & 0.23 & 0.15 & 0.08 & 0.03 & 0.03 & 0.22 & 0.14 & 0.08 & 0.04 & 0.04\\ 
        RetinexNet & 0.66 & 0.56 & 0.66 & 0.67 & 0.71 & 0.44 & 0.42 & 0.64 & 0.57 & 0.54 & 0.39 & 0.45 & 0.40 & 0.53 & 0.47\\
        LIME & 0.28 & 0.19 & 0.12 & 0.08 & 0.07 & 0.19 & 0.11 & 0.06 & 0.04 & 0.04 & 0.23 & 0.14 & 0.08 & 0.05 & 0.04\\
        UNet & 0.86 & 0.84 & \bf{0.83} & 0.76 & 0.75 & 0.78 & 0.75 & 0.71 & 0.65 & \bf{0.63} & 0.82 & 0.77 & \bf{0.73} & 0.67 & 0.65\\ 
        VST+VBM4D & 0.79 & 0.72 & 0.57 & 0.34 & 0.30 & 0.63 & 0.51 & 0.25 & 0.13 & 0.11 & 0.53 & 0.41 & 0.30 & 0.15 & 0.14\\ 
        FastDVDNet & 0.74 & 0.78 & 0.67 & 0.56 & 0.52 & 0.65 & 0.65 & 0.56 & 0.43 & 0.44 & 0.60 & 0.47 & 0.25 & 0.32 & 0.31\\ 
        KPN & 0.75 & 0.73 & 0.61 & 0.49 & 0.46 & 0.74 & 0.67 & 0.20 & 0.14 & 0.13 & 0.74 & 0.63 & 0.47 & 0.34 & 0.33\\ 
        MBLLVEN & 0.80 & 0.76 & 0.75 & 0.70 & 0.61 & 0.66 & 0.65 & 0.60 & 0.53 & 0.37 & 0.77 & 0.73 & 0.67 & 0.60 & 0.58\\
        SMOID & 0.85 & 0.84 & 0.82 & 0.76 & 0.75 & 0.75 & 0.73 & 0.70 & 0.63 & 0.60 & 0.81 & 0.76 & 0.70 & 0.64 & 0.66\\
        TC-ResUNet & \bf{0.87} & \bf{0.85} & 0.83 & \bf{0.77} & \bf{0.76} & \bf{0.78} & \bf{0.76} & \bf{0.72} & \bf{0.65} & 0.63 & \bf{0.83} & \bf{0.78} & 0.72 & \bf{0.68} & \bf{0.66} \\
        \midrule
    \end{tabular}}
    \end{center}
    \label{tab:enh_results_dynamic}
\end{table*}

\vspace{1mm}
\noindent\textbf{Visual Results.~~~~}
The exemplar visual results of the restoration algorithms are shown in Figs.~\ref{fig:result_img_rec} and \ref{fig:result_vid_rec}. Scaling converts a visually unpleasant dark image/frame into a bright one with amplified noise. After applying BM3D, the image is smoothed and noise is partly removed but still exists. LIME enhances a low-light image/frame by estimating its illumination map and reflectance. With our extremely dark data, LIME's results are still dark and noisy, so we apply BM3D and scaling to its original outputs. The final results display color distortion and obvious noise. RetinexNet produces dark and noisy results with bad color and under-enhanced areas (black regions). VST+VBM4D results in more smoothness in the restored image than BM3D by effective Poisson noise denoising and additionally considering temporal consistency, but the resulted image also suffers from severe color distortion that is caused by VST. Despite the fact that FastDVDNet and KPN retain good color consistency, residual noise appears in the restored images. MBLLVEN produces a result with color distortion. UNet and SMOID produce visually pleasant results for most of the cases. Noise is efficiently removed and details are reconstructed without color distortion problem. However, for the high-end camera data, SMOID produces shadow-like artifacts near the roadblock cone. In comparison, the proposed TC-ResUNet also produces good visual results, with better noise suppression than UNet. The zoom-in regions in Fig.~\ref{fig:result_vid_rec} validate that our method leads to fewer reconstruction artifacts and better image quality.

The exemplar visual results at different TRs are shown in Figs.~\ref{fig:result_rec_comp_STAR} and \ref{fig:result_rec_comp_STAR_vid}.
Visual quality degrades with the decrease of TR. The pleasant reconstruction with less challenging input validates the progress in the past years in denoising and low-light image/video restoration, and the severe artifacts in low TR data results reveal the great challenges of restoring the low-light data under severe noise and extremely weak ambient illumination. Therefore, more efforts are expected in developing techniques/algorithms to handle highly noisy input at extreme scenarios.
We leave further analysis and comparisons among different methods to the following quantitative evaluation. 

\vspace{1mm}
\noindent\textbf{Quantitative analysis.~~~~}
The quantitative evaluation of the reconstruction algorithms is shown in Tables \ref{tab:enh_results_static} and~\ref{tab:enh_results_dynamic}. The trend of performance change with TR and camera sensitivity consists with the visual results. 

These results indicate that as TR becomes lower, restoration difficulty grows drastically and performance deteriorates greatly. When TR decreases to a certain level, applying restoration algorithms would not help that much in recovering the details immersed in noise. When TR falls from the highest to lowest, the best reconstruction PSNR(dB)/SSIM of low-end camera changes from 29.56/0.85 to 26.66/0.78 (static image) and from 27.39/0.83 to 18.35/0.66 (dynamic video), and with middle-end and high-end camera data the best PSNR/SSIM also shows an obvious decrease. The great challenge calls for novel powerful algorithms to address such low-illuminance cases.
Besides, there exists a large difference in the imaging quality and reconstruction results among different cameras. One thing to be noted is, because the three cameras possess different responses in pixel intensity, the absolute values are not comparable. For example, low-end camera has the strongest response (highest pixel value) and the baseline methods result in largest reconstruction error. 

From the tables, we can also observe that as for conventional noise removal and enhancement methods, 
the performance of BM3D and LIME are both superior to linear scaling. Comparatively, deep-learning methods lead to obviously higher PSNR/SSIM. UNet consistently achieves best results across 5 groups of data among all the existing algorithms for the image subset. Benefiting from its superior ability in feature representation and prior modeling, deep-learning based methods work better than conventional image restoration algorithms on low-light data.

\begin{table*}[t!]
    \centering
    \caption{The statistical significance of performance difference between TC-ResUNet and UNet tested on DRV, SDSD and DarkVision.}
    \vspace{-2mm}
    \begin{center}
    \resizebox{1\textwidth}{!}{
    \begin{tabular}{l|c|c|c|c}
        \midrule
        Dataset & \multicolumn{4}{c}{P-value of PSNR / P-value of SSIM}\\
        \toprule
        DRV & \multicolumn{4}{c}{0.000074222402 /0.000000000000}\\
        \midrule
        SDSD & \multicolumn{4}{c}{0.000024239992/ 0.001178737344}\\
        \midrule
        \multirow{7}{*}{DarkVision} & TR (\%) & High-end camera	& Middle-end camera & Low-end camera \\
        \midrule
        & 3.2 & 0.000000000000 / 0.000000000000 & 0.000000000000 / 0.000000000000 & 0.000000000000 / 0.000000000000 \\
        & 1.6 & 0.000011919401 / 0.000000000000 & 0.000000000000 / 0.000000000000 & 0.000000000000 / 0.000000000000 \\
        & 0.8 & 0.000016843342 / 0.000000000000	& 0.000000000000 / 0.000000000002 & 0.000003674504 / 0.000000000717 \\
        & 0.4 & 0.000000041511 / 0.000000000000 & 0.000000000006 / 0.000000000002 & 0.000000000059 / 0.000000000000 \\
        & 0.2 & 0.000000000000 / 0.000000000000	& 0.000000000000 / 0.000000000000 & 0.000000321379 / 0.000000396550 \\   
        \midrule
    \end{tabular}}
    \end{center}
    \label{tab:p_value}
\end{table*}

\begin{table*}[t!]
    \centering
    \caption{The comparison of object detection performance with and without applying enhancement on dark static images.}
    \vspace{-2mm}
    \begin{center}
    \resizebox{1\textwidth}{!}{
    \begin{tabular}{l|rrrrr|rrrrr|rrrrr}
        \midrule
        & \multicolumn{5}{c|}{High-end camera} & \multicolumn{5}{c|}{Middle-end camera} & \multicolumn{5}{c}{Low-end camera}\\
        TR (\%) & 3.2 & 1.6 & 0.8 & 0.4 & 0.2 & 3.2 & 1.6 & 0.8 & 0.4 & 0.2 & 3.2 & 1.6 & 0.8 & 0.4 & 0.2 \\
        \toprule
        w/o enhancement & 58.5 & 56.9 & 42.5 & 29.7 & 31.5 & 47.1 & 34.5 & 20.2 & 9.4 & 10.2 & 35.7 & 30.3 & 19.1 & 12.8 & 10.2\\
        w/ enhancement & 63.8 & 60.7 & 47.8 & 40.4 & 38.8 & 46.2 & 44.1 & 32.3 & 19.6 & 18.6 & 45.5 & 42.5 & 34.9 & 22.1 & 21.2 \\
        mAP improvement & +5.2 & +3.8 & +5.3 & +10.7 & +7.3 & -0.9 & +9.6 & +12.1 & +10.2 & +8.4 & +9.8 & +12.3 & +15.8 & +9.3 & +11.0 \\
        \midrule
    \end{tabular}}
    \end{center}
    \label{tab:cross_task_results_detection_static}
\end{table*}

\begin{table*}[t!]
    \centering
    \caption{The comparison of object detection performance with and without applying enhancement on dark video frames.}
    \vspace{-2mm}
    \begin{center}
    \resizebox{1\textwidth}{!}{
    \begin{tabular}{l|rrrrr|rrrrr|rrrrr}
        \midrule
        & \multicolumn{5}{c|}{High-end camera} & \multicolumn{5}{c|}{Middle-end camera} & \multicolumn{5}{c}{Low-end camera}\\
        TR (\%) & 3.2 & 1.6 & 0.8 & 0.4 & 0.2 & 3.2 & 1.6 & 0.8 & 0.4 & 0.2 & 3.2 & 1.6 & 0.8 & 0.4 & 0.2 \\
        \toprule
        w/o enhancement & 55.7 & 48.9 & 36.3 & 28.3 & 13.8 & 54.1 & 41.9 & 29.3 & 9.6 & 4.8 & 43.5 & 34.8 & 26.6 & 7.9 & 5.2\\
        w/ enhancement & 66.0 & 58.8 & 54.9 & 32.2 & 19.8 & 68.8 & 50.3 & 45.0 & 19.1 & 11.8 & 63.4 & 46.9 & 32.3 & 16.0 & 20.0 \\
        mAP improvement & +10.2 & +9.8 & +18.6 & +3.9 & +5.9 & +14.7 & +8.4 & +15.7 & +9.5 & +7.0 & +19.9 & +12.1 & +5.7 & +8.1 & +14.9 \\
        \midrule
    \end{tabular}}
    \end{center}
    \label{tab:cross_task_results_detection_video}
\end{table*}

\begin{table*}[t!]
    \centering
    \caption{The comparison between generic and category-specific (vehicle in this figure) enhancement on dark videos, respectively with randomly cropped patches or bounding-box cropped ones as training data.}
    \vspace{-2mm}
    \begin{center}
    \resizebox{1\textwidth}{!}{
    \begin{tabular}{l|rrrrr|rrrrr|rrrrr}
        \midrule
        \bf{PSNR (dB)} & \multicolumn{5}{c|}{High-end camera} & \multicolumn{5}{c|}{Middle-end camera} & \multicolumn{5}{c}{Low-end camera}\\
        TR (\%) & 3.2 & 1.6 & 0.8 & 0.4 & 0.2 & 3.2 & 1.6 & 0.8 & 0.4 & 0.2 & 3.2 & 1.6 & 0.8 & 0.4 & 0.2 \\
        \toprule
        Random cropping & 25.14 & 23.62 & 23.24 & 20.94 & 20.53 & 25.97 & 25.16 & 21.76 & 18.49 & 17.50 & 25.55 & 22.13 & 20.83 & 18.04 & 18.27\\
        Bounding-box cropping & 25.82 & 24.05 & 23.74 & 21.18 & 20.04 & 26.50 & 25.04 & 21.62 & 18.27 & 17.46 & 26.01 & 22.47 & 21.02 & 17.85 & 18.11 \\
        PSNR improvement & +0.68 & +0.43 & +0.51 & +0.24 & -0.48 & +0.53 & -0.12 & -0.14 & -0.22 & -0.03 & +0.47 & +0.34 & +0.19 & -0.19 & -0.16 \\
        \midrule
        \bf{SSIM} & \multicolumn{5}{c|}{High-end camera} & \multicolumn{5}{c|}{Middle-end camera} & \multicolumn{5}{c}{Low-end camera}\\
        TR (\%) & 3.2 & 1.6 & 0.8 & 0.4 & 0.2 & 3.2 & 1.6 & 0.8 & 0.4 & 0.2 & 3.2 & 1.6 & 0.8 & 0.4 & 0.2 \\
        \toprule
        Random cropping & 0.85 & 0.83 & 0.78 & 0.73 & 0.72 & 0.80 & 0.74 & 0.70 & 0.63 & 0.63 & 0.81 & 0.74 & 0.71 & 0.63 & 0.65\\
        Bounding-box cropping & 0.86 & 0.84 & 0.80 & 0.74 & 0.71 & 0.80 & 0.75 & 0.70 & 0.63 & 0.62 & 0.82 & 0.75 & 0.72 & 0.63 & 0.65 \\
        SSIM improvement & +0.01 & +0.01 & +0.02 & +0.01 & -0.01 & +0.00 & +0.01 & +0.00 & -0.00 & -0.01 & +0.01 & +0.01 & +0.01 & -0.00 & -0.00 \\
        \midrule
    \end{tabular}}
    \end{center}
    \label{tab:cross_task_results_enhancement}
\end{table*}

On the video benchmark, VST+VBM4D results in higher PSNR/SSIM than other conventional iterative methods. This improvement is mainly attributed to effective Poisson noise denoising for low-light data and exploitation of temporal information. Deep-learning based methods show diverse performances. Methods such as RetinexNet, KPN and MBLLVEN may fail utterly in certain extreme cases. SMOID leads to good performances consistently. Among these methods, TC-ResUNet achives the best results for most cases. The statistical significance of performance difference between TC-ResUNet and UNet is also tested, as shown in Table~\ref{tab:p_value}. These results imply that (1) encoder-decoder structure is effective in low-light data denoising and enhancement and (2) multiple frames contribute more information to the network, so it can learn better representation of original image and noise distribution. In Table~\ref{tab:p_value} the higher significance level (lower P-value) on DarkVision reveals a more solid conclusion on DarkVision than on previous dark video datasets (DRV and SDSD). 
Besides, such systematical analysis of the performances of different algorithms conveys a message that a good trade-off among camera sensitivity, algorithm complexity and applicable illuminance level is quite important in real applications. 
\subsection{Interplay between high-level and low-level vision tasks}
In spite that there are some preliminary studies on jointly conducting image/video reconstruction and high-level tasks under low illumination, the interplay between low-level and high-level dark vision (i.e., image/video restoration and object detection) remains an open question to be explored. Here we design some experiments to study the interaction in two reverse directions: the assistance of image/video restoration for object detection, and the guidance from object detection for restoration.

\vspace{1mm}
\noindent\textbf{From low-level task towards high-level task.~~~~}
To test the benefits brought by image restoration for object detection, we apply image restoration algorithms to enhance the low-light image/video before object detection, and compare the results with those by direct detection. In this group of experiment, we adopt UNet/TC-UNet for restoration and Faster R-CNN for image/video object detection. Specifically, we use well-trained restoration models to generate enhanced images/videos, which are used to train the above mentioned detectors to make them more suitable for the enhanced data.

The results are shown in Tables.~\ref{tab:cross_task_results_detection_static} and~\ref{tab:cross_task_results_detection_video}, from which we can observe that restoring the low-light data will consistently help detection across different illuminance levels among all cameras on both static and dynamic data. The performance improvement is more significant on the low quality data, from lower-end cameras and at lower TRs. Quantitatively, on the low-end camera data the accuracy improvement at 5 illuminance levels ranges from 9.3 to 15.8 (static image) and from 5.7 to 19.9, and the largest improvement is obtained at TR 0.8\% and 3.2\% respectively. On the contrary, applying restoration beforehand does not help quite a lot on the high TR data by the high-end camera, because the recorded data itself is of high quality and contains abundant discriminative features for detection. 
Overall, these results indicate that applying enhancement is helpful for high-level tasks on low-light data, especially when severely degenerated by photon starvation.

\vspace{1mm}
\noindent\textbf{From high-level task towards low-level task.~~~~}
In order to explore the guidance from object detection to image enhancement, we perform restoration on (1) randomly cropped areas and (2) areas containing certain category objects. For the second model, we implement object detector to obtain predicted bounding boxes and then crop these areas. For fair comparison, We make the two models learn from the same size of training dataset in terms of pixel number.

The results shown in Table~\ref{tab:cross_task_results_enhancement} tell that at higher TRs, class-specific restoration models tend to perform better than general models, and such class-specific models can act as a successive ROI enhancer after object detection; at lower TRs, however, this advantage is less apparent or even turns disadvantage, since detection results in these cases are unreliable. These observations demonstrate that in low-light scenarios, ROI training strategy is helpful in enhancing the object categories of interest if provided with reliable bounding boxes. Such enhancement strategy is of lower computation cost, lower model complexity and higher performance on ROIs. Besides, the object detection under extremely dark environment still demands further research efforts and progress.

\subsection{From DarkVision to real low-light image}
\begin{figure}[t]
	\centering
	\begin{minipage}[ht]{0.999\textwidth}
		\subfigure[]{
			\includegraphics[width=0.235\textwidth]{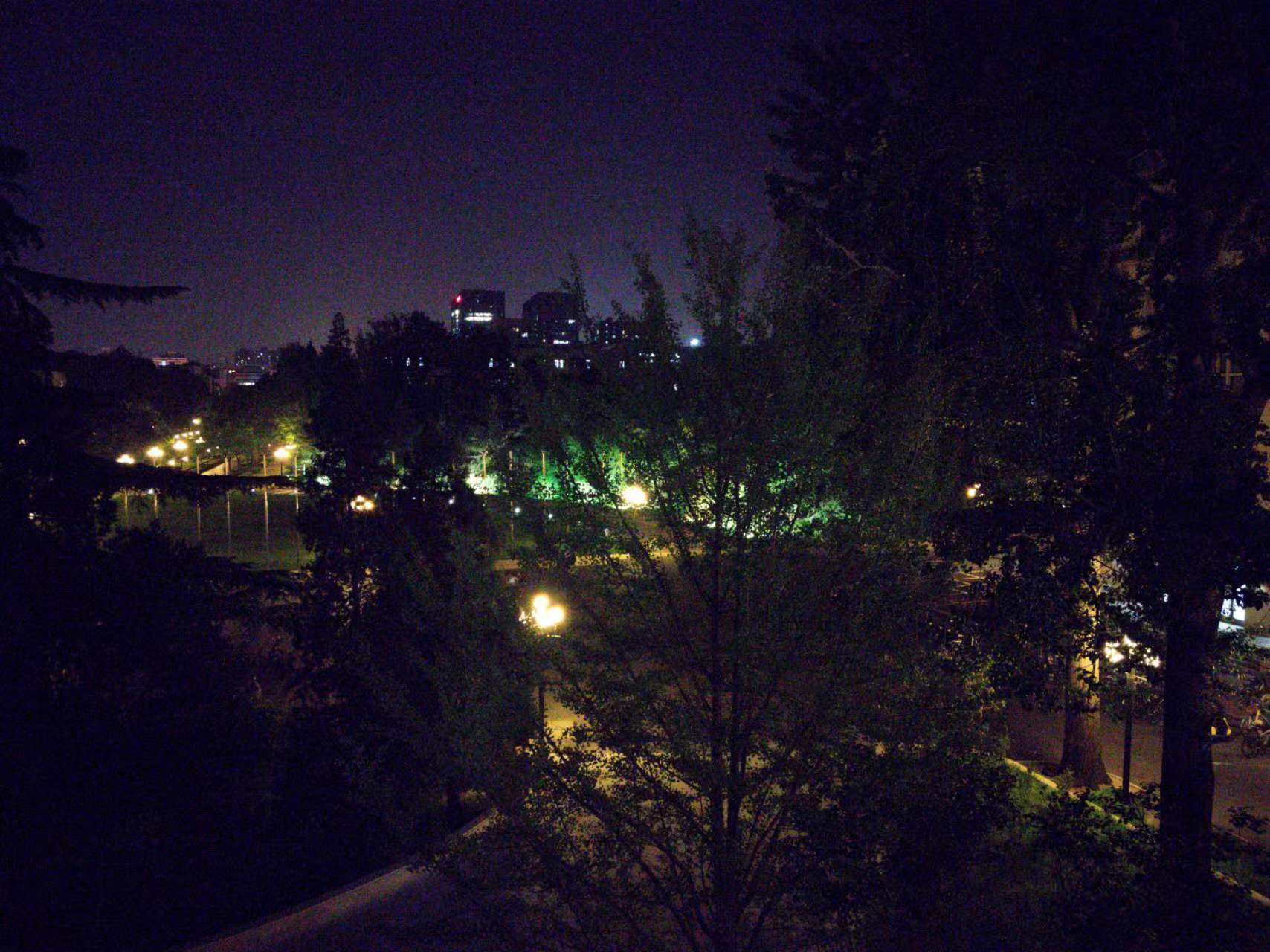}
		}
		\subfigure[]{
			\includegraphics[width=0.235\textwidth]{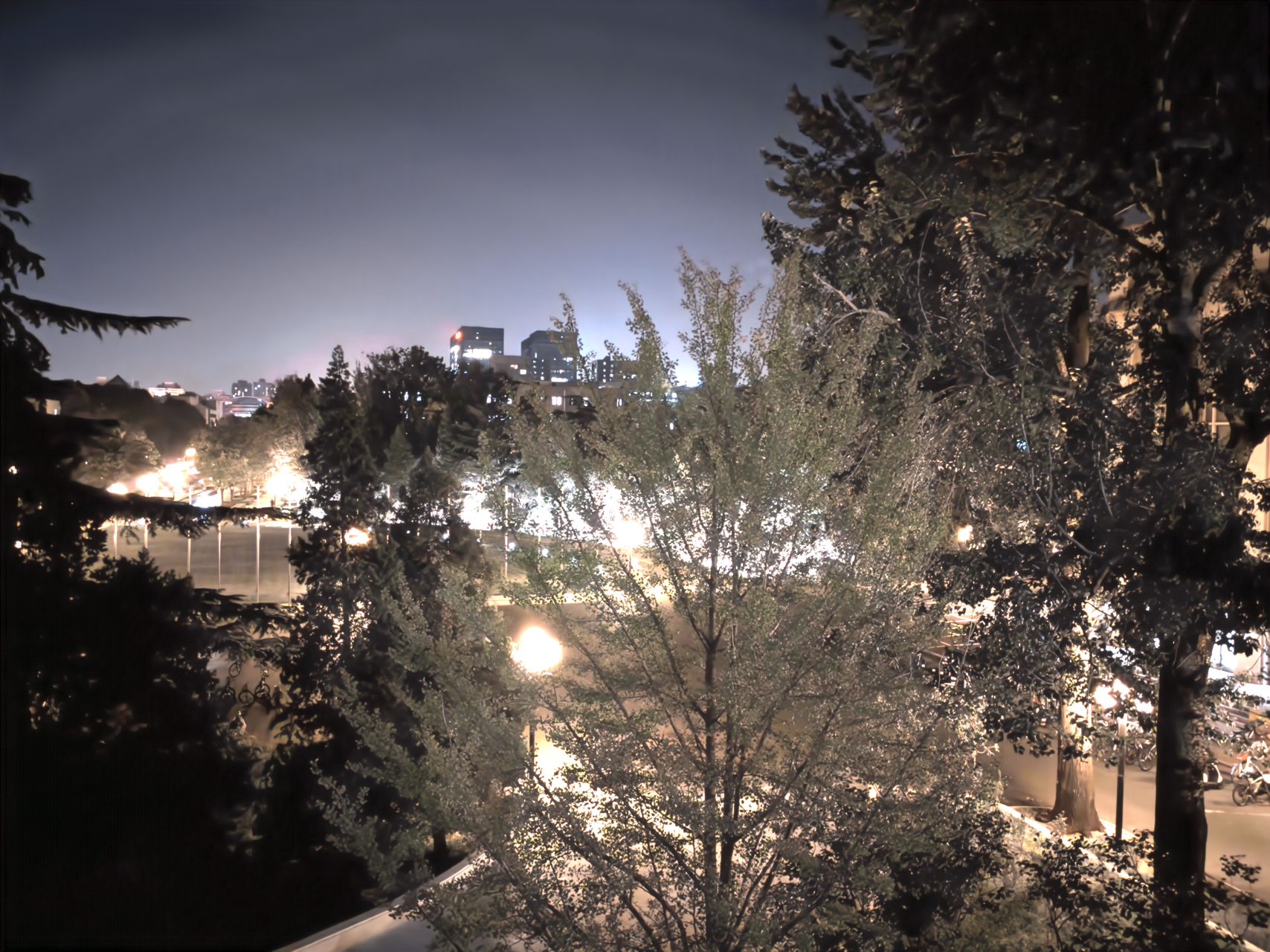}
		}
	\end{minipage}
	\caption{(a) A real nighttime photo taken by Xiaomi 11 smartphone with auto-enhancement. (b) Restoration result by the model trained on DarkVision.}
	\label{fig:real}
\end{figure}

When capturing the dataset, we managed to collect diverse scenes with varying illuminations since there exists difference between real dark (e.g. nighttime) images and those by attenuating a normal-light image with ND filters, e.g., the recording times vary from dawn to dusk and places include both indoors and outdoors.
To validate the generalization of the models trained from DarkVision on real low-light images, we take a real nighttime photo and feed it to the network well-trained from the closest illumination level. Here we capture nighttime photos with a Xiaomi 11 smartphone with exposure time set to 500 ms and ISO 1600. We remove black level from the raw data, demosaic the Bayer images, scale the pixel values to the JAI 3.2 subset mean intensity, and predict an enhanced color frame from the TC-ResUNet model. One typical example is shown in Fig.~\ref{fig:real} (a), for which we obtain a plausible enhanced frame in (b) that restores rich details with good color fidelity and fine details. This result shows that serving as a work-around strategy, our DarkVision has made a step closer to handling actual low-light images.

\section{Summary and Discussions}\label{sec:conclusion}
In this paper we have introduced DarkVision, a new large-scale dataset for low-light data analysis. By means of a two-path acquisition setup, the DarkVision dataset contains both static and dynamic data suitable for image/video enhancement and object detection tasks. In order to establish a benchmark, we have evaluated various approaches for both tasks on the DarkVision. In addition, we have also explored the temporal consistency in low-light videos. The experiment results have revealed the great challenges of DarkVision and demonstrated the effectiveness of our methods. It has also been shown that the DarkVision can contribute to the cross-task analysis for low-light data.

The studies would push forward researches in surveillance at night, night photography, astronomical photography, etc. Low-light imaging is also highly related to other fields, such as high-speed imaging which also suffers from low photon flux due to short exposure. We hope this database would serve as a benchmark for testing the performance of different low-light methods, studying the quantitative/theoretical analysis on low photon number imaging, open new research directions and benefit real-life applications such as autonomous driving, biomedical microscopy, fluorescence imaging.


\section*{Acknowledgment}

This work is jointly funded by Ministry of Science and Technology of China (Grant No. 2020AAA0108202), National Natural Science Foundation of China (Grant No. 61931012 and 62088102) and Beijing Natural Science Foundation (Grant No. Z200021).

\section*{Data Availability}
The datasets generated during and/or analysed during the current study are available from the corresponding author on reasonable request.

\bibliographystyle{spbasic}
\bibliography{main}


\end{document}